\documentclass[pdflatex,sn-mathphys-num]{sn-jnl}

\usepackage{graphicx}
\usepackage{multirow}
\usepackage{amsmath,amssymb,amsfonts}
\usepackage{amsthm}
\usepackage{mathrsfs}
\usepackage[title]{appendix}
\usepackage{xcolor}
\usepackage{textcomp}
\usepackage{manyfoot}
\usepackage{booktabs}
\usepackage{algorithm}
\usepackage{algorithmicx}
\usepackage{algpseudocode}
\usepackage{listings}
\usepackage{subfig}

\newcommand\ringring[1]{%
  {
   \mathop{\kern0pt #1}\limits^{
     \vbox to-1.85ex{
       \kern-2ex 
       \hbox to 0pt{\hss\normalfont\kern.1em \r{}\kern-.45em \r{}\hss}%
       \vss 
     }
   }
  }
}

\newcommand\ringringring[1]{%
  {
   \mathop{\kern0pt #1}\limits^{
     \vbox to-1.85ex{
       \kern-2ex 
       \hbox to 0pt{\hss\normalfont\kern.1em \r{}\kern-.45em \r{}\kern-.45em \r{}\hss}%
       \vss 
     }
   }
  }
}

\begin{document}

\title[A Game Between Two Identical Dubins Cars]{A Game Between Two Identical Dubins Cars:  Evading a Conic Sensor in Minimum Time}


\author[]{\fnm{Ubaldo} \sur{Ruiz}}\email{uruiz@cicese.mx}

\affil[]{\orgname{Centro de Investigaci\'on Cient\'ifica y de Educaci\'on Superior de Ensenada (CICESE)}, \orgaddress{\city{Ensenada}, \postcode{22860}, \state{Baja California}, \country{M\'exico}}}


\abstract{A fundamental task in mobile robotics is keeping an intelligent agent under surveillance with an autonomous robot as it travels in the environment. This work studies a theoretical version of that problem involving one of the most popular vehicle platforms in robotics. In particular, we consider two identical Dubins cars moving on a plane without obstacles. One of them plays as the pursuer, and it is equipped with a limited field-of-view detection region modeled as a semi-infinite cone with its apex at the pursuer's position. The pursuer aims to maintain the other Dubins car, which plays as the evader, as much time as possible inside its detection region. On the contrary, the evader wants to escape as soon as possible. In this work, employing differential game theory, we find the time-optimal motion strategies near the game's end. The analysis of those trajectories reveals the existence of at least two singular surfaces: a Transition Surface (also known as a Switch Surface) and an Evader's Universal  Surface. We also found that the barrier's standard construction produces a surface that partially lies outside the playing space.}




\maketitle

\section{Introduction}

A fundamental task in mobile robotics is keeping an intelligent agent under surveillance with an autonomous robot as it travels in the environment. This task can be modeled as a pursuit-evasion game with two players having antagonistic goals. In this work, we study a theoretical version of a surveillance pursuit-evasion problem considering one of the most commonly used vehicle models in robotics, a Dubins car. In our problem, two identical Dubins cars move on a plane without obstacles. One plays as the pursuer, aiming to keep the other player inside its limited field of view. Having an opposite goal, the evader seeks to escape from it as soon as possible. The pursuer's detection region is modeled as a semi-infinite cone with its apex at the pursuer's position. 

We model the problem as a zero-sum differential game. In particular, by performing a retro-time integration starting from the game's terminal conditions, we compute the players' motion strategies near the end of it. The analysis of the corresponding trajectories reveals the existence of at least two singular surfaces: a Transition Surface (also known as a Switch Surface) and an Evader's Universal Surface. Additionally, by performing the standard barrier's construction to solve the problem of deciding the game's winner, we found that it produces a surface that partially lies outside the playing space.

As was mentioned, we model our problem as a zero-sum differential game. Rufus Isaacs \cite{ISAACS-99} developed a methodology to solve differential games, which is employed in this paper. The fundamental idea is partitioning the playing space into regions where the value function is differentiable. Usually, the process's most challenging part is identifying the regions' boundaries, called {\em singular surfaces}. Characterizing a singular surface and its outcome is frequently based on the premise that one player must base his control choice on the knowledge of his opponent's control selection. A strategy computed employing this information is called a non-admissible strategy. In contrast, an admissible strategy does not demand further information on the players' controls and is established only on knowledge of the system's state. In our work, we succeed in finding mathematical equations describing the players' time-optimal motion strategies near the game's end and reaching two types of singular surfaces. Refer to \cite{ISAACS-99,BASAR-98,LEWIN-12}, for a detailed study of Isaacs' methodology and singular surfaces. 

This article is organized as follows. The related work and contributions are presented in Section \ref{sec:prev}, and the problem definition is introduced in Section \ref{sec:problemdefinition}. The set of configurations where the game ends is computed in Section \ref{sec:terminalcond}. In Section \ref{sec:motionstrategies}, the players' time-optimal motion strategies near the game's end are obtained. In Section \ref{sec:barrierfirst}, an attempt to solve the decision problem is described, and it is exhibited that it produces a barrier surface that partially lies outside the playing space. A brief discussion about the players' motion strategies and the corresponding trajectories found in this work is presented in Section \ref{sec:summary}. Numerical simulations illustrating the players' motion strategies are shown in Section \ref{sec:simulation}, and the conclusions and future work are described in Section \ref{sec:conclusions}.

\section{Related work}
\label{sec:prev}

This paper studies a pursuit-evasion game \cite{MERZ-72}. In the literature, many works have addressed pursuit-evasion games \cite{ISAACS-99,BASAR-98,LEWIN-12,MERZ-74}. Usually, they are grouped into three main categories: search \cite{GUIBAS-99,TOVAR-08,HOLLINGER-09}, capture \cite{MERZ-72,KARNAD-09,RUIZ-13,LI-16,BERA-17,BRAVO-20,YAN-2022,RUIZ-22b,CHAUDHARI-21}, and tracking \cite{GERKEY-06,BHATTACHARYA-10,MURRIETA-11,BECERRA-16,STIFFLER-17,ZOU-18,RUIZ-22,LOZANO-22,LOZANO-22b,RUIZ-24}. In the first category, the pursuer's objective is to find the evader while both players move in an environment with obstacles, i.e., put the evader inside the pursuer's detection region. In the second category, the pursuer strives to capture the evader by attaining a certain distance from it. In this category, usually, the capture wants to be achieved as soon as possible. In the third category, the pursuer's goal is to keep surveillance of the evader as both players advance in the environment, i.e., maintain the evader inside the pursuer's detection region. The problem addressed in this paper belongs to the last category. For a more detailed taxonomy of pursuit-evasion problems, we refer the reader to the following surveys \cite{CHUNG-11,ROBIN-16}. In the next paragraphs, we summarize and briefly compare the works most related to this paper in differential games' literature. To the best of our knowledge, we believe those works are \cite{MERZ-72,BERA-17,CHAUDHARI-21,RUIZ-23,GREENFELD-1987,LEWIN-79}.

In \cite{MERZ-72}, a pursuit-evasion problem involving two identical Dubins Cars is analyzed. The pursuer's goal is to capture the evader in minimum time. The evader, on the contrary, wants to avoid the pursuer as much time as possible. First, the game of kind is solved, i.e., determine the states from which capture is feasible. Second, for those states where capture is possible, the game of degree is solved, i.e., the optimal controls of the players to achieve their goals are found. More recently, \cite{BERA-17} presents a comprehensive solution to that problem, \cite{CHAUDHARI-21} provides feedback-based solutions for particular cases, and \cite{BUZIKOV-23} presents an exhaustive analytical description of the barrier for all values of the capture radius.

In \cite{RUIZ-24}, the pursuit-evasion game of surveillance evasion between two identical Differential Drive Robots is studied. In that problem, one of the Differential Drive Robots plays as the pursuer, and it is equipped with a bounded range sensor modeled as a circle centered at the pursuer's location. Similar to our current work, the pursuer's objective is to maintain surveillance of the evader as much as possible while the evader seeks to escape as soon as possible. In that work, the problem of deciding the game's winner, i.e., whether the evader escapes or not, is solved. Additionally, the players' time-optimal strategies when the evader escapes surveillance are provided. Note that our problem differs from \cite{RUIZ-24} in two ways: 1) the players are Dubins cars, which have different time-optimal motion primitives than a Differential Drive Robot, and 2) in our case, the sensor is a semi-infinite cone and not a circle. Those two changes result in the motion strategies and the conditions deciding the game's winner not being the same, requiring a completely new analysis.

In \cite{RUIZ-23}, the problem of keeping surveillance of an Omnidirectional Agent with a Differential Drive Robot equipped with a limited field of view sensor is analyzed. Similar to our current work, the sensor is modeled as a semi-infinite cone fixed to the Differential Drive Robot's body. However, since the pursuer and the evader have different kinematic constraints than a Dubins car, the players' motion strategies in that work differ from those found in our current work. Additionally, dealing with two non-holonomic players requires using a higher dimensional space representation, which makes the analysis harder to perform. 

A differential game of surveillance between two identical Dubins cars was studied in \cite{GREENFELD-1987}. That work presents a partial solution to the problem where the pursuer is equipped with a circular detection region. Like our work, the pursuer wants to keep the evader inside its detection region as much as possible, while the evader has the opposite goal. However, different from \cite{GREENFELD-1987}, the pursuer has a semi-infinite conic detection region in our game. It is important to stress that this change directly impacts the evader's escape condition. In \cite{GREENFELD-1987}, the escape is attained when the evader reaches a certain distance from the pursuer, while, in our current work, it is accomplished when the relative orientation of the evader from the pursuer's location is greater than the angle defining the semi-infinite cone. That may seem like a minor difference; however, it has been systematically observed in differential games' literature that altering the sensor's constraints requires a new problem analysis since the players' motion strategies to achieve their goals and the singular surfaces appearing in the game change. 

In \cite{LEWIN-79}, another pursuit-evasion game of surveillance is analyzed. In that work, a Dubins car pursuer wants to maintain an Omnidirectional Agent inside its detection region. Like our work, the pursuer is equipped with a limited field of view sensor modeled as a semi-infinite cone. However, since the evader in that case is an Omnidirectional Agent, the players' motion strategies differ from the ones found in our current work. As was pointed out before, having two Dubins cars as players also implies requiring a higher dimensional space representation of the problem than the one in \cite{LEWIN-79}, which makes finding a solution a more difficult problem.

\subsection{Contributions}

The main contributions of this work are:
\begin{itemize}
    \item We compute the players' time-optimal motion strategies near the game's end. The corresponding trajectories are described by analytical expressions.
    \item We reveal the existence of two singular surfaces: a Transition Surface, where one of the players switches its control, and an Evader's Universal Surface. We also found the players' motion strategies and the corresponding trajectories that reach those surfaces.
    \item We exhibit that the usual procedure of constructing the barrier from the boundary of the usable part produces a surface that partially lies outside the playing space.
\end{itemize}

\section{Problem definition}
\label{sec:problemdefinition}

\begin{figure}[tbh]
\centering
\subfloat[Realistic space \label{fig:realistic}]{
\includegraphics[width=0.4\linewidth]{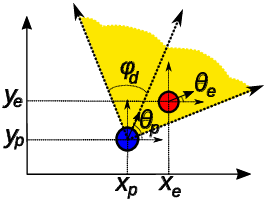}
}
\subfloat[Reduced space \label{fig:reduced}]{
\includegraphics[width=0.4\linewidth]{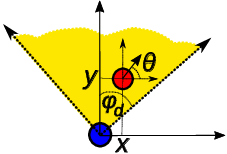}
}
\caption{The pursuer and the evader are represented by the blue and red dots, respectively.} 
\end{figure}

Two identical Dubins cars with unit speed and unit turn radius move on a plane without obstacles. One of them plays as the {\em pursuer}, and it is equipped with a limited field-of-view (FoV) detection region modeled as a semi-infinite cone with its apex at the pursuer's position. The pursuer aims to maintain the other Dubins car, which plays as the {\em evader}, as much time as possible inside its detection region. On the contrary, the evader wants to escape as soon as possible. The pursuer's FoV is modeled as a semi-infinite cone with half-angle $\phi_d\in(0,\frac{\pi}{2})$ fixed to its location and aligned with its heading (see Fig. \ref{fig:realistic}). In this work, only kinematic constraints are considered. 

We employ two representations to analyze and display the player's motion strategies. In the first one, which is known as {\em the realistic space} and we use Cartesian coordinates, $(x_p,y_p,\theta_p)$ represents the pursuer's pose, and $(x_e,y_e,\theta_e)$ represents the evader's pose (see Fig. \ref{fig:realistic}). Thus, the state of the system can be denoted as $\mathbf{x}=(x_p,y_p,\theta_p,x_e,y_e,\theta_e)\in\mathbb{R}^2\times S^1\times \mathbb{R}^2 \times S^1$. The following equations describe the players' motions in the realistic space
\begin{equation}
\begin{split}
\dot{x}_p &= \cos \theta_p, \:\:\:\:\: \dot{y}_p = \sin \theta_p, \:\:\:\:\: \dot{\theta}_p = \nu_p,  \\
\dot{x}_e &= \cos \theta_e, \:\:\:\:\: \dot{y}_e = \sin \theta_e, \:\:\:\:\: \dot{\theta}_e = \nu_e, \\
\end{split}
\end{equation}
where $\nu_p\in [-1,1]$ is the pursuer's control and $\nu_e\in[-1,1]$ corresponds to the evader's control. All angles are measured counter-clockwise from the positive $x$-axis in this representation. 

In the second representation, we employ a coordinate transformation in which the reference frame is fixed to the pursuer's location, and the $y$-axis is aligned with its motion direction (see Fig. \ref{fig:reduced}). All angles are measured in a clockwise direction from the $y$-axis. This representation is known as {\em the reduced space}, and it is obtained using the following coordinate transformation
\begin{equation}
\label{eq:coordinatetrans}
\begin{split}
    x &= (x_e-x_p) \sin \theta_p - (y_e-y_p) \cos \theta_p, \\
	y &= (x_e-x_p) \cos \theta_p + (y_e-y_p) \sin \theta_p, \\
    \theta &= \theta_p - \theta_e.
\end{split}    
\end{equation}
The system's state in the reduced space is denoted as $\mathbf{x}_R=(x,y,\theta)$. Computing the time derivative of (\ref{eq:coordinatetrans}), we get the following kinematic equations
\begin{equation}
\label{eq:reducedsystem}
\begin{split}
\dot{x} &= \nu_p y + \sin \theta, \:\:
\dot{y} = -\nu_p x - 1 + \cos \theta, \:\:
\dot{\theta} = \nu_p  - \nu_e,
\end{split}
\end{equation}
where again $\nu_e,\nu_p\in[-1,1]$ denote the pursuer's and the evader's controls, respectively. They can be expressed as $\dot {\mathbf{x}}_R = f(\mathbf{x}_R,\nu_e,\nu_p)$. Having a cylindrical representation of the state $\mathbf{x}_c=(r,\phi,\theta)$ in the reduced space is also convenient. Here, $r$ is the distance from the origin to the evader's location, $\phi$ is the angle between the pursuer's heading ($y$-axis) and the evader's location, and $\theta$ is defined in the same way as in (\ref{eq:coordinatetrans}). The kinematic equations of the cylindrical representation are given by
\begin{equation}
\label{eq:polarsystem}
\begin{split}
\dot r &= \cos(\theta - \phi) - \cos \phi, \:\:
\dot \phi = \nu_p  + \frac{\sin(\theta-\phi)+\sin \phi}{r}, \:\:
\dot{\theta} = \nu_p  - \nu_e.
\end{split}
\end{equation}
In this work, during the analysis of the problem, we indistinctly switch between the two coordinate representations of the state in the reduced space.

In Fig. \ref{fig:reduced}, we can observe that the conic detection region is split into two symmetric parts by the $y$-axis. In the following paragraphs, we mainly describe the construction of motion strategies in the right part of the semi-infinite cone, i.e., $\phi \in [0,\phi_d]$. The trajectories in the left part, i.e., $\phi\in[-\phi_d,0]$ can be obtained employing some symmetries around the $y$-axis. Occasionally, we will make reference to the left part of the cone and its trajectories to describe some interesting behaviors we have found in this game.

\section{Terminal conditions}
\label{sec:terminalcond}

One fundamental step in solving a differential game is to find the initial conditions used to perform the retro-time integration of the motion equations \cite{ISAACS-99, BASAR-98,LEWIN-12}. In our game, those configurations where the evader is located at the boundary of the conic detection region and can escape regardless of the controls applied by the pursuer are used as initial conditions, and they are called the {\em usable part} (UP). In this game, the UP is composed of two sets, one corresponding to the right boundary of the cone $(\phi=\phi_d)$ and the other to the left boundary $(\phi=-\phi_d)$. We denote the portion of the right boundary as the {\em right usable part} (RUP); on it, the evader can increase the value of $\phi$ regardless of the controls applied by the pursuer. The following equation represents the previous condition
\begin{equation}
\label{eq:UPa}
\mbox{RUP}=\Bigg\{(r,\phi_d,\theta):\min_{\nu_p} \max_{\nu_e} \dot \phi > 0\Bigg\}, 
\end{equation}
or 
\begin{equation}
\label{eq:UPb}
\mbox{RUP}=\Bigg\{(r,\phi_d,\theta):\min_{\nu_e} \max_{\nu_p} - \dot \phi < 0\Bigg\}, 
\end{equation}
to follow the convention in the problem definition that the pursuer is the maximizer player, and the evader is the minimizer player. Substituting (\ref{eq:polarsystem}) into (\ref{eq:UPb}), we get
\begin{equation}
\label{eq:UPmax}
\begin{split}
\mbox{RUP}=\Bigg\{(r,\phi_d,\theta):&\min_{\nu_e} \max_{\nu_p} \Bigg[ -\nu_p - \frac{\sin(\theta-\phi_d)+\sin \phi_d}{r} <0 \Bigg] \Bigg\}. 
\end{split}
\end{equation}
For $\phi_d\in(0,\frac{\pi}{2})$, we have that $
\nu_p=-1$ maximizes (\ref{eq:UPmax}). Also, note that $\dot \phi$ is independent of the value of $\nu_e$. Substituting $\nu_p=-1$ into (\ref{eq:UPmax}) we get
\begin{equation}
\label{eq:UP1}
\begin{split}
\mbox{RUP}=\Bigg\{(r,\phi_d,\theta):1 - \frac{\sin(\theta-\phi_d)+\sin \phi_d}{r} < 0 \Bigg\}. 
\end{split}
\end{equation}
By doing some algebraic manipulation, we found that
\begin{equation}
\mbox{RUP}=\Bigg\{(r,\phi_d,\theta):r<\sin(\theta-\phi_d)+\sin\phi_d \Bigg\}.
\end{equation}
Using a similar approach, we found the configurations where the evader is located at the left boundary of the conic detection region and can decrease the value of $\phi$ regardless of the controls applied by the pursuer escaping from the detection region. In this case, we have that $\nu_p=1$. We denote the set of those configurations as the {\em left usable part} (LUP)
\begin{equation}
\mbox{LUP}=\Bigg\{(r,\phi_d,\theta):r<-\sin(\theta+\phi_d)+\sin\phi_d \Bigg\}.
\end{equation}
A portion of the line formed by the intersection of the semi-infinite planes defining the detection region at $r=0$ belongs to the RUP. We denote it as RUPL, and corresponds to the following configurations
\begin{equation}
\label{eq:RUPL}
\mbox{RUPL}=\Bigg\{(0,\phi_d,\theta):0<\sin(\theta-\phi_d)+\sin\phi_d \Bigg\}.
\end{equation}
Similarly, a portion of the line formed by the intersection of the semi-infinite planes defining the detection region at $r=0$ belongs to the LUP. We denote it as LUPL, and corresponds to the following configurations
\begin{equation}
\label{eq:LUPL}
\mbox{LUPL}=\Bigg\{(0,\phi_d,\theta):0<-\sin(\theta+\phi_d)+\sin\phi_d \Bigg\}.
\end{equation}
In this game, we found that RUPL and LUPL intersect for $\phi_d\in(0,\frac{\pi}{2})$, and as $\phi_d \rightarrow \frac{\pi}{2}$, RUPL $\simeq$ LUPL. The {\em boundary of the usable part} (BUP) corresponds to the configurations where neither of the players unilaterally increase or decrease the value of $\phi$, i.e.,
\begin{equation}
\label{eq:BUPeq}
\mbox{BUP}=\Bigg\{(r,\phi_d,\theta):\min_{\nu_e} \max_{\nu_p} \dot \phi = 0\Bigg\}, 
\end{equation}
in our game, those configurations are given by
\begin{equation}
\label{eq:RBUP}
\mbox{RBUP}=\Bigg\{(r,\phi_d,\theta):r=\sin(\theta-\phi_d)+\sin\phi_d \Bigg\}.
\end{equation}
for the right boundary of the cone, and 
\begin{equation}
\label{eq:LBUP}
\mbox{LBUP}=\Bigg\{(r,\phi_d,\theta):r=-\sin(\theta+\phi_d)+\sin\phi_d \Bigg\}.
\end{equation}
for the left one.
From (\ref{eq:RBUP}), we have that, $r=0$ at $\theta=0$ and $\theta=\pi+2\phi_d$, and $r=1 + \sin \phi_d$ (maximum value) at $\theta=\frac{\pi}{2}+\phi_d$. Note that the RUP contains only configurations where $\theta\in[0,\pi+2\phi_d]$, one can verify that the values of $\theta \in (\pi+2\phi_d, 2\pi)$ correspond to negative values of $r$, which are invalid. Since $r=0$ for $\theta=0$ and $\theta=\pi+2\phi_d$, those configurations in the RBUP are part of the line formed by the intersection between the planes defining the detection region, and we denoted them as RBUPL. Similarly, from (\ref{eq:LBUP}), in the LUP we have that $r=0$ for $\theta=\pi-2\phi_d$ and $\theta=2\pi$, and $r=1+\sin \phi_d$ for $\theta=\frac{3\pi}{2}-\phi_d$. In this case, the configurations where $r=0$ are denoted as LBUPL.

\begin{figure}[tp]
\centering
\subfloat[Right Usable Part (RUP) \label{fig:RUP}]{
\includegraphics[scale=0.2]{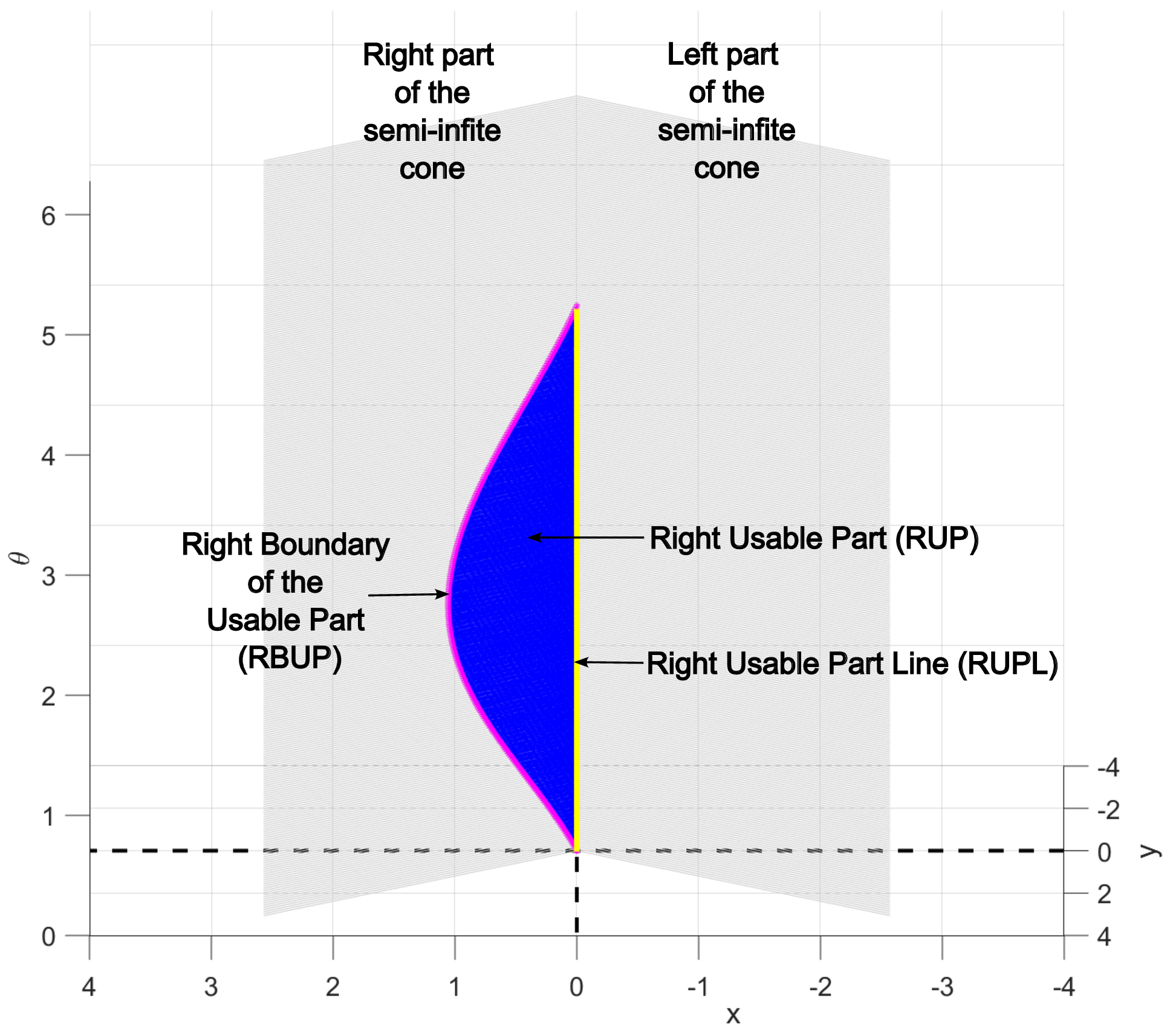}
}
\subfloat[Left Usable Part (LUP) \label{fig:LUP}]{
\includegraphics[scale=0.2]{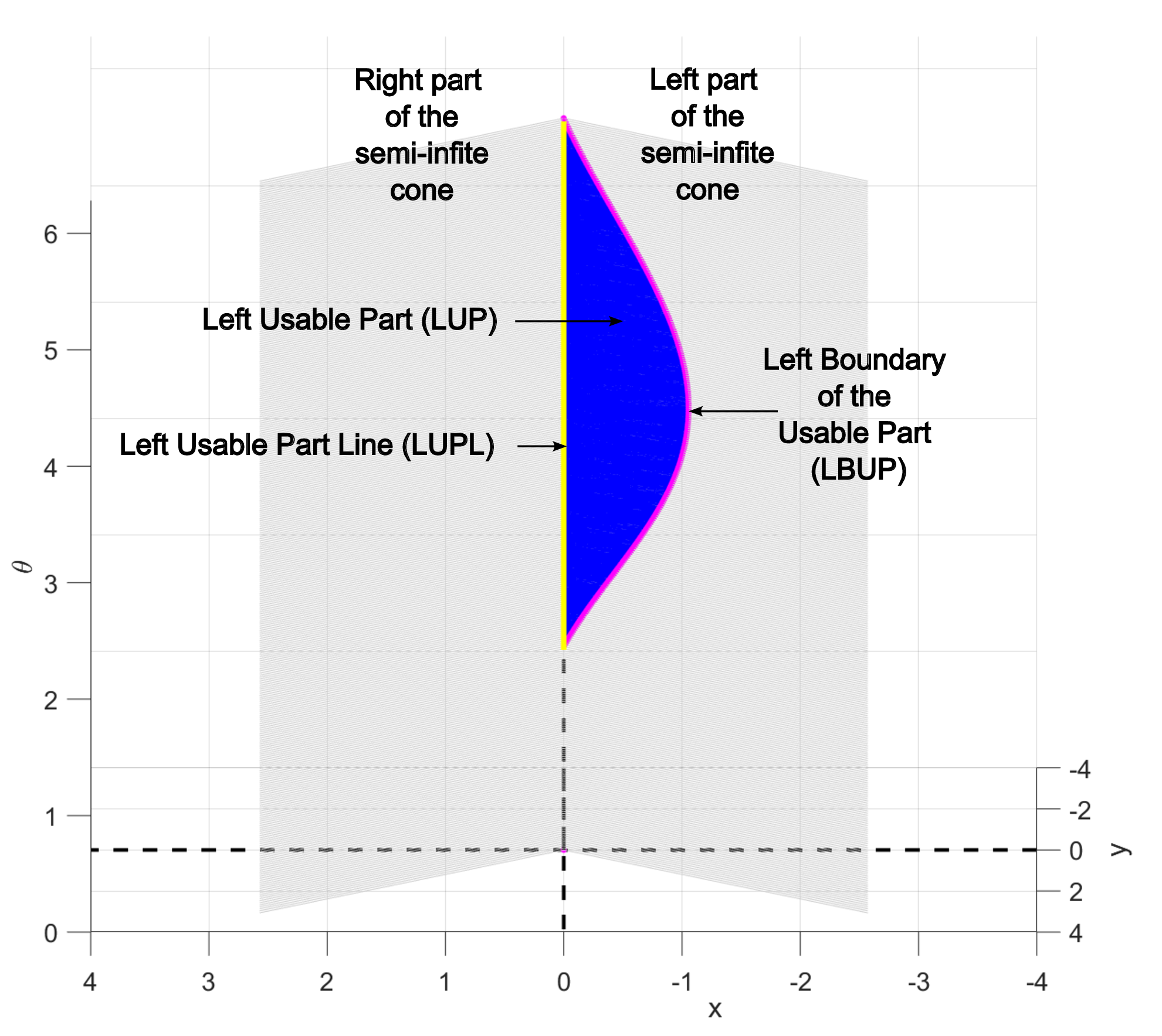}
}
\hfill
\subfloat[Usable Part (UP) and its boundary (BUP)]{
\includegraphics[scale=0.34]{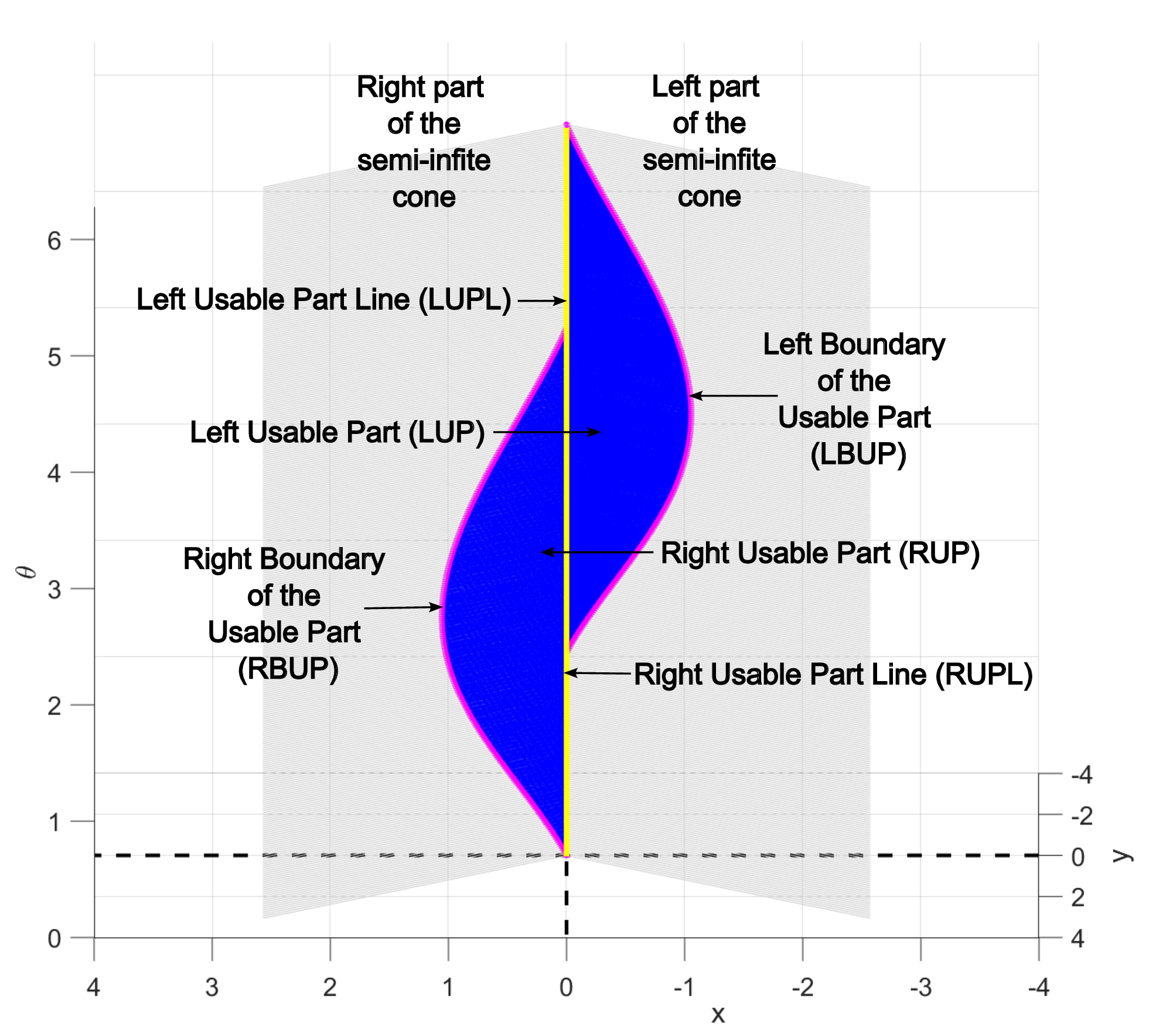}
}
\caption{Representation of the Usable Part (UP), blue region, and its boundary (BUP), magenta curves, in the reduced space for $\phi_d=40^\circ$. The Usable Part Line (UPL) corresponds to the yellow line. The gray rectangles represent the semi-infinite cone as $\theta$ varies from $0$ to $2\pi$. The image shows views from an observer located in front of the semi-infinite cone.}
\label{fig:UP}
\end{figure}

Fig. \ref{fig:UP} shows a representation of the UP and the BUP, for $\phi_d=40^\circ$. In that figure, it is possible to see a very interesting behavior that we have systematically observed for $\phi_d\in(0,\frac{\pi}{2})$. In particular, we refer to the line formed by the intersection of the semi-infinite planes defining the detection region. From (\ref{eq:RUPL}) and (\ref{eq:RBUP}), we have that the portion of that line corresponding to $\theta\in(0,\pi+2\phi_d)$ belongs to the RUPL, and from (\ref{eq:LUPL}) and (\ref{eq:LBUP}), the portion of that line corresponding to $\theta\in(\pi-2\phi_d,2\pi)$ belongs to the LUPL. Note that the RUPL and LUPL intersect for $\theta\in (\pi-2\phi_d,\pi+2\phi_d)$, and in this case, the evader escapes by either crossing the right or left boundary. Also, note that for $\theta\in(0,\pi-2\phi_d)$, since that portion belongs to the RUPL, the evader can escape by crossing the right boundary; however, if the evader focuses on crossing the left boundary of the detection region it will not succeed. An analogous behavior occurs for the portion of the line where $\theta\in(\pi+2\phi_d, 2\pi)$. This indicates that in the last two cases, escape occurs; however, the evader needs to focus on the correct boundary and apply the corresponding control. Note that $\theta=\pi-2\phi_d$ corresponds to a configuration in the LBUPL, but it also belongs to the RUPL; thus, despite the evader cannot escape by focusing on the left boundary, it can accomplish that goal if it focuses on the right boundary. A similar behavior occurs for $\theta=\pi+2\phi_d$, which belongs to the RBUPL and the LUPL. Later in the paper, we focus on solving the decision problem, and we analyze the case $\theta=0$ ($\theta=2\pi$), which exhibits an interesting behavior.

\section{Motion strategies}
\label{sec:motionstrategies}

In this section, we compute the players' optimal strategies to attain their goals in the right portion of the semi-infinite cone, i.e., $\phi\in[0,\phi_d]$. Following the methodology described in \cite{ISAACS-99,BASAR-98,LEWIN-12}, a retro-time integration of the players' motion equations is performed, taking the configurations at the UP as initial conditions. In the following, we denote the retro-time as $\tau=t_f-t$, where $t_f$ is the termination time of the game.

\subsection{Optimal controls}

First, we need to find the expressions of the optimal controls used by the players during the game. To do that, we have to construct the Hamiltonian of the system. From \cite{BASAR-98}, we have that
\begin{equation}
\label{eq:Hamiltonian}
H(\mathbf{x},\mathbf{\lambda},\nu_e,\nu_p) = \mathbf{\lambda}^T \cdot f(\mathbf{x},\nu_e,\nu_p) + L(\mathbf{x},\nu_e,\nu_p),
\end{equation}
where $\lambda^T$ are the costate variables and $L(\mathbf{x},\nu_e,\nu_p)$ is the cost function. Recalling that $L(\mathbf{x},\nu_e,\nu_p)=1$ for problems of minimum time \cite{ISAACS-99}, like the one addressed in this paper, and substituting (\ref{eq:reducedsystem}) into (\ref{eq:Hamiltonian}), we have that in the reduced space and Cartesian coordinates
\begin{equation}
\label{eq:hamiltoniancartesian}
\begin{split}
H(\mathbf{x}, \lambda, \nu_p,\nu_e) &= \lambda_{x} \nu_p y  + \lambda_{x} \sin \theta - \lambda_{y}\nu_p  x - \lambda_{y} + \lambda_{y} \cos \theta + \lambda_{\theta} \nu_p  -\lambda_{\theta} \nu_e+1. \\
\end{split}
\end{equation}
The optimal controls are obtained from (\ref{eq:hamiltoniancartesian}) and  Pontryagin's Maximum Principle, which states that along the systems' optimal trajectories
\begin{equation}
\label{eq:hamiltonianmax}
\begin{split}
\min_{\nu_e} \max_{\nu_p} H(\mathbf{x}, \lambda, \nu_e, \nu_p)=0,&\\
\nu_e^*=\arg \min_{\nu_e} H(\mathbf{x}, \lambda, \nu_e, \nu_p), &\:\:
\nu_p^*=\arg \max_{\nu_p} H(\mathbf{x}, \lambda, \nu_e, \nu_p),\\
\end{split}
\end{equation}
where $\nu_p^*$ and $\nu_e^*$ denote the optimal controls of the pursuer and the evader, respectively. Thus, we have that the pursuer's optimal control in the reduced space and Cartesian coordinates is given by
\begin{equation}
\label{eq:pursuerctrls}
\nu_p^* =  \mbox{sgn}\left(y\lambda_x -x\lambda_y + \lambda_\theta \right),
\end{equation}
and if we define $S=y\lambda_x -x\lambda_y + \lambda_\theta$, (\ref{eq:pursuerctrls}) can be rewritten as
\begin{equation}
\label{eq:pursuerctrls2}
\nu_p^* =  \mbox{sgn}\left(S \right).
\end{equation}
The evader's optimal control is given by
\begin{equation}
\label{eq:evaderctrls}
\nu_e^* = \mbox{sgn}\left(\lambda_\theta\right).
\end{equation}

\subsection{Costate equation}
From (\ref{eq:pursuerctrls}) and (\ref{eq:evaderctrls}), one can notice that for computing the players' optimal controls, we require to know the values of $\lambda^T=[\lambda_x \:\: \lambda_y \:\: \lambda_{\theta}]^T$ as time elapses. To find those values, we use the costate equation. In particular, since we are going to perform a retro-time integration of the motion equations, we have to use the retro-time version of the costate equation
\begin{equation}
\label{eq:costate}
\mathring \lambda =\frac{\partial}{\partial x}H(\mathbf{x}, \lambda, \nu_e^*, \nu_p^*),
\end{equation}
where the retro-time derivative of a variable $x$ is represented is by $\mathring x$. Substituting (\ref{eq:reducedsystem}) into (\ref{eq:costate}), and considering the players' optimal controls $\nu_e^*$ and $\nu_p^*$ we have that
\begin{equation}
\label{eq:adjointeq}
\begin{split}
&\mathring \lambda_{x}= -\nu_p^*\lambda_{y}, \:\: \mathring \lambda_{y}= \nu_p^*\lambda_{x}, \:\:\mathring \lambda_{\theta} = \lambda_x \cos \theta - \lambda_y \sin \theta.  \\
\end{split}
\end{equation}
We need to find the initial conditions at the game's end ($\tau = 0$) to perform the retro-time integration of (\ref{eq:adjointeq}). At the UP, $x=r\sin\phi_d$, $y=r\cos\phi_d$ and $\theta=\theta_d$ (the final difference between the players' headings). From the traversability conditions, we have that
\begin{equation}
\label{eq:adjointinitial}
\lambda_{x} = -\cos \phi_d, \:\: \lambda_{y} = \sin \phi_d, \:\: \lambda_{\theta} = 0.
\end{equation}
Integrating (\ref{eq:adjointeq}) considering the initial conditions in (\ref{eq:adjointinitial}), we have that
\begin{equation}
\label{eq:adjointsolutions}
\begin{split}
\lambda_{x} &= -\cos\left(\phi_d - \nu_p^*\tau \right), \:\:
\lambda_{y} = \sin\left(\phi_d -  \nu_p^*\tau \right),\\
\lambda_\theta &= \nu_e^*\left(-\sin(\phi_d-\theta_d) + \sin(\phi_d-\theta_d- \nu_e^*\tau)\right).
\end{split}
\end{equation}

\subsection{Primary solution}
\label{sec:primarysol}

\begin{figure}[t]
\centering
\includegraphics[scale=0.35]{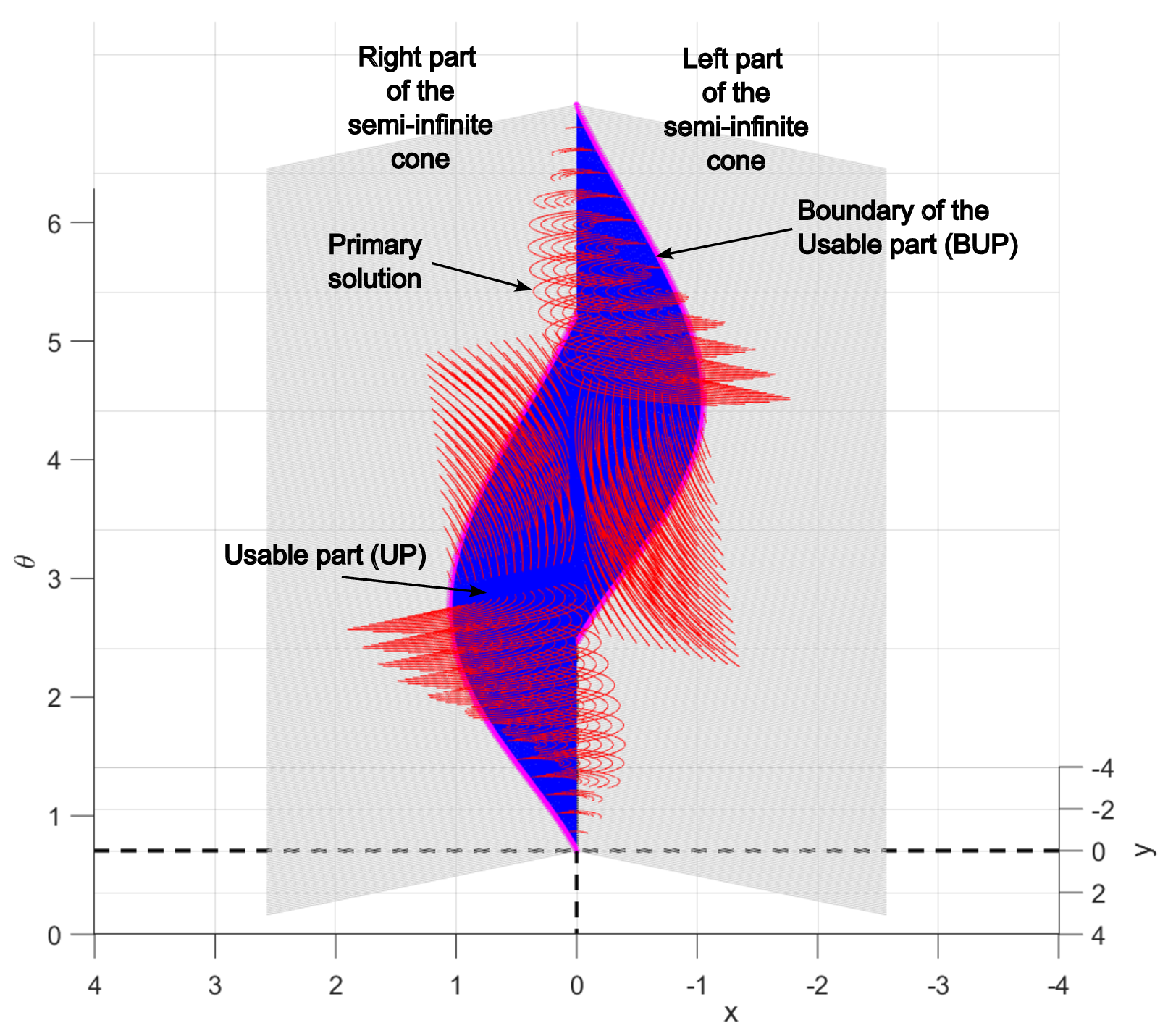}
\caption{Primary solution (red curves) for $\phi_d=40^\circ$. The trajectories start from the UP and continue backward in time until they hit the boundary of the detection region (gray region) again, or the pursuer switches its control. Note that for all configurations in the gray region $\phi=\phi_d$, however, the evader cannot immediately escape, i.e., they do not belong to the UP.
}
\label{fig:primary_sol}
\end{figure}

Now, we compute the trajectories of the players that lead directly to the terminal conditions (see Fig. \ref{fig:primary_sol}). Those trajectories are known as the {\em primary solution}. To compute them, we need the retro-time version of the motion equations 
\begin{equation}
\label{eq:retroreducedsystem}
\begin{split}
\mathring{x} &= -\nu_p y - \sin \theta, \:\:
\mathring{y} = \nu_p x  + 1 - \cos \theta, \:\:
\mathring{\theta} = -\nu_p + \nu_e.
\end{split}
\end{equation}
Integrating (\ref{eq:retroreducedsystem}), considering the initial conditions $x=r\sin\phi_d$, $ y=r\cos\phi_d$ and $\theta=\theta_d$ at the UP, and the players' optimal controls $\nu_p^*$ and $\nu_e^*$, we have that
\begin{equation}
\label{eq:primary}
\begin{split}
x &= -\nu^*_p+{\nu^*_p}\cos\left(\nu^*_p \tau \right)-\nu_e^*\cos\left(\theta_d-\nu^*_p \tau\right)+\nu_e^*\cos\left(\theta_d+\left(\nu^*_e - \nu^*_p\right)\tau\right) \\
&+r\sin\left(\phi_d- \nu^*_p\tau\right),\\
y &= \nu_p^*\sin\left(\nu^*_p\tau\right)+r\cos\left(\phi_d- \nu^*_p\tau\right)-2\nu_e^*\cos\left(\theta_d+\left(\frac{\nu^*_e}{2}-\nu^*_p\right)\tau\right)\sin\left(\frac{\nu^*_e}{2}\tau\right),\\
\theta &= \theta_d+\left(\nu^*_e-\nu^*_p \right)\tau.
\end{split}
\end{equation}

This solution is valid as long as the players do not switch controls. In this game, we found that after following the trajectories in (\ref{eq:primary}) for some time, the pursuer switches its control. To find the time when that change occurs, we need to substitute (\ref{eq:adjointsolutions}) and (\ref{eq:primary}) into (\ref{eq:pursuerctrls}), and compute the value of $\tau$ (the retro-time variable) take makes the resulting expression equal to zero. However, given (\ref{eq:primary}) contains transcendental expressions involving $\tau$, the result of the previous substitution is also a transcendental equation. Thus, it is not possible to directly estimate $\tau$ and numerical methods are required to determine its value. Later in the paper, we discuss how to compute the trajectories after the pursuer's control switch.

During the simulations, we also found that some of the retro-time primary trajectories depart from the boundary of the detection region, and after some time, they meet that boundary again before the pursuer’s switch occurs.  Unfortunately, to validate analytically under which conditions this behavior occurs, we are forced again to face transcendental equations given the nature of (\ref{eq:primary}), limiting our efforts in that direction.

Two examples of trajectories appearing in the primary solution and the corresponding motion of the players in the realistic space are presented in Section \ref{sec:simulation}.

\subsection{Discussion of the previous solution}

Note that $\lambda_{\theta}=0$ at the game's end, then from (\ref{eq:evaderctrls}), $\nu_e^*=0$. Thus, we must check whether the evader continues using that control or switches it immediately. We can do that by analyzing the value of $\mathring \lambda_{\theta}$ at the game's end. 

Substituting the terminal conditions $\lambda_x=-\cos \phi_d$, $\lambda_y=\sin \phi_d$ and $\theta=\theta_d$ into (\ref{eq:adjointeq}), we have that
\begin{equation}
\label{eq:lambdathfinal}
\mathring \lambda_{\theta} = -\cos(\phi_d-\theta_d).
\end{equation}
From (\ref{eq:lambdathfinal}), we can deduce that $\mathring \lambda_{\theta}\neq 0$ for any value of $\phi_d-\theta_d\neq \pm \frac{\pi}{2}$. This implies that in those cases, we can use the sign of $\mathring \lambda_{\theta}$ to compute the control of $\nu_e^*$ immediately before the game's end and substitute its value in (\ref{eq:retroreducedsystem}) and (\ref{eq:primary}).

For $\phi_d-\theta_d = \pm \frac{\pi}{2}$, we can check whether $\ringring \lambda_{\theta}$ is different from zero. From (\ref{eq:adjointeq}), and recalling again that $\lambda_x=-\cos \phi_d$, $\lambda_y=\sin \phi_d$ and $\theta=\theta_d$ at the game's end, we found that
\begin{equation}
\ringring \lambda_{\theta} = 0.
\end{equation}
where the second order retro-time derivative of $\lambda_{\theta}$ is represented is by $\ringring \lambda_{\theta}$.
As described in previous works \cite{ISAACS-99,MERZ-72,GREENFELD-1987}, this suggests the existence of an Evader's Universal Surface (EUS). On that surface, the evader applies $\nu_e^*=0$, which can be verified using Isaacs' necessary condition for the existence of Universal Surfaces \cite{ISAACS-99}. 

\subsection{Evader's Universal Surface and its tributary trajectories}

In this section, we construct the trajectories corresponding to the EUS (see Fig. \ref{fig:universal_surface}). Recalling that $v_e^*=0$, we have the retro-time version of the motion equations take the form
\begin{equation}
\label{eq:retroreducedsystemUS}
\begin{split}
\mathring{x} &= -\nu_p y - \sin \theta, \:\: \mathring{y} = \nu_p x  + 1 - \cos \theta, \:\:
\mathring{\theta} = -\nu_p.
\end{split}
\end{equation}

Integrating (\ref{eq:retroreducedsystemUS}), considering the initial conditions $x=r\sin\phi_d$, $y=r\cos \phi_d$ and $\theta_d=\phi_d-\frac{\pi}{2}$ at the UP, and the players' optimal controls, we get
\begin{equation}
\label{eq:US}
\begin{split}
x &= -\nu^*_p+\nu^*_p\cos\left(\nu^*_p \tau \right)+r\sin\left(\phi_d- \nu^*_p\tau\right)-\tau\sin\left(\theta_d- \nu^*_p\tau\right),\\
y &= \nu_p^*\sin\left(\nu^*_p\tau\right)+r\cos\left(\phi_d- \nu^*_p\tau\right)-\tau\cos\left(\theta_d- \nu^*_p\tau\right),\\
\theta &= \theta_d-\nu^*_p\tau.
\end{split}
\end{equation}
From the traversability conditions, in this case, we have that
\begin{equation}
\label{eq:adjointinitialUS}
\lambda_{x} = -\cos \phi_d, \:\: \lambda_{y} = \sin \phi_d, \:\: \lambda_{\theta} = 0.
\end{equation}
Integrating (\ref{eq:adjointeq}) considering the initial conditions in (\ref{eq:adjointinitialUS}), we get
\begin{equation}
\label{eq:adjointsolutionsUS}
\begin{split}
\lambda_{x} &= -\cos\left(\phi_d - \nu_p^*\tau \right), \:\:
\lambda_{y} = \sin\left(\phi_d - \nu_p^*\tau \right), \:\:
\lambda_\theta = 0.
\end{split}
\end{equation}

\subsubsection{Tributary trajectories}

\begin{figure}[t]
    \centering
    \includegraphics[scale=0.35]{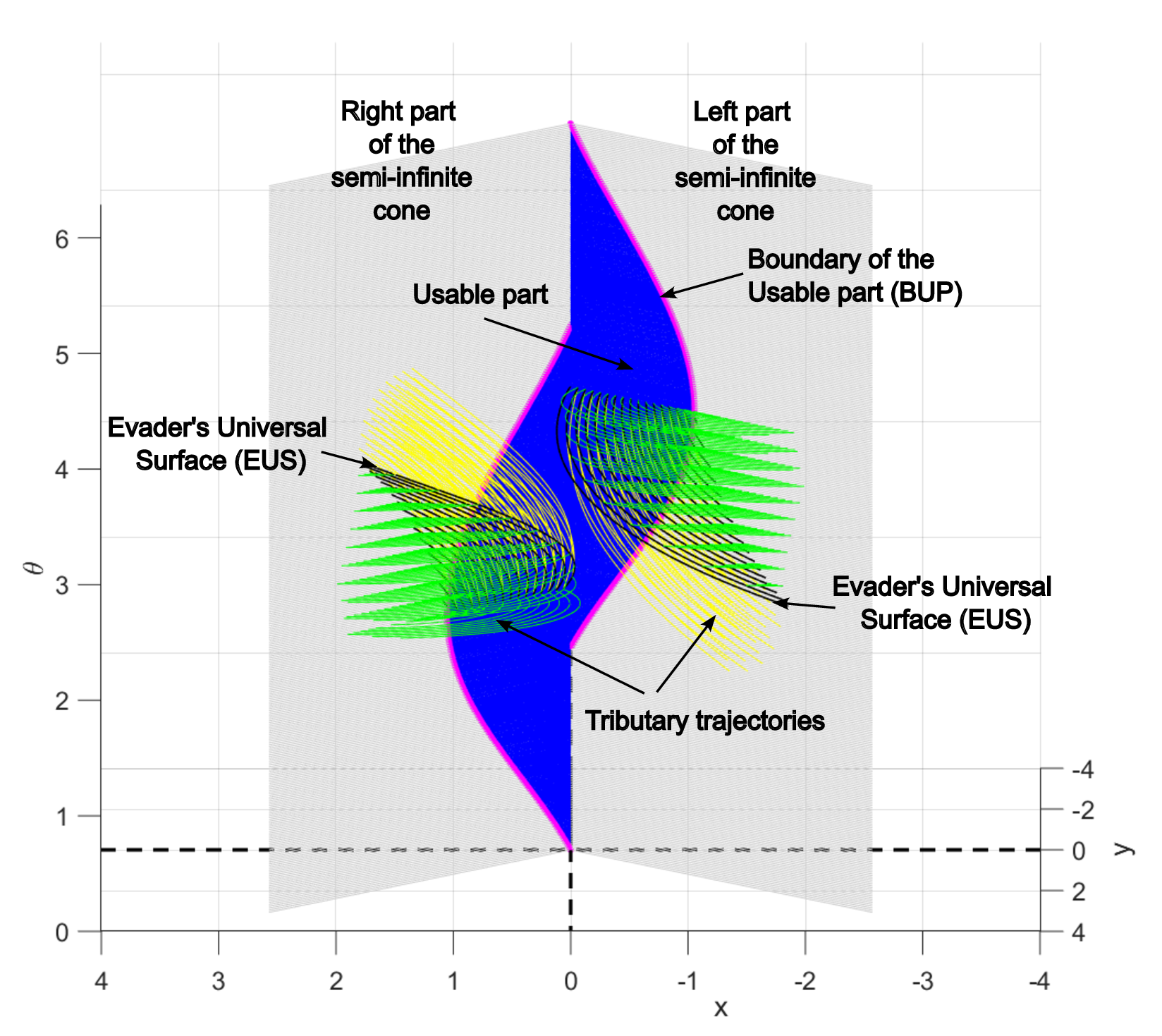}
    \caption{An example of the Evader's Universal Surface (black curves) and its tributary trajectories for $\phi_d=40^\circ$. The green and yellow colors indicate the trajectories at each side of the EUS. The evader applies a particular control at each side, i.e., $\nu_e^*=-1$ or $\nu_e^*=1$. In retro-time, the tributary trajectories continue until they hit the boundary of the detection region or the pursuer switches its control.}
    \label{fig:universal_surface}
\end{figure}

Now, we compute the tributary trajectories reaching the EUS (see Fig. \ref{fig:universal_surface}). In this case, the retro-time version of the motion equations is
\begin{equation}
\label{eq:retroreducedsystemtributary}
\begin{split}
\mathring{x} &= -\nu_p y - \sin \theta, \:\:
\mathring{y} = \nu_p x  + 1 - \cos \theta, \:\:
\mathring{\theta} = -\nu_p + \nu_e.
\end{split}
\end{equation}
Integrating (\ref{eq:retroreducedsystemtributary}), considering as initial conditions the configurations $(x_{US},y_{US},\theta_{US})$ at the EUS, and the players' optimal controls, we have that
\begin{equation}
\small
\label{eq:tributaryUS}
\begin{split}
x &= -\nu^*_p+\nu^*_p\cos\left(\nu^*_p (\tau-\tau_{US}) \right)-\nu_e^*\cos\left(\theta_{US}-\nu^*_p (\tau-\tau_{US})\right)\\
&+\nu_e^*\cos\left(\theta_{US}+\left(\nu^*_e - \nu^*_p\right)(\tau-\tau_{US})\right)+r_{US}\sin\left(\phi_{US}- \nu^*_p(\tau-\tau_{US})\right),\\
y &= \nu_p^*\sin\left(\nu^*_p(\tau-\tau_{US})\right)+r_{US}\cos\left(\phi_{US}- \nu^*_p(\tau-\tau_{US})\right)\\
&-2\nu_e^*\cos\left(\theta_{US}+\left(\frac{\nu^*_e}{2}-\nu^*_p\right)(\tau-\tau_{US})\right)\sin\left(\frac{\nu^*_e}{2}(\tau-\tau_{US})\right),\\
\theta &= \theta_{US}+\left(\nu^*_e-\nu^*_p \right)(\tau-\tau_{US}),
\end{split}
\end{equation}
where $(r_{US},\phi_{US},\theta_{US})$ are the cylindrical coordinates of $(x_{US},y_{US},\theta_{US})$ and $\tau_{US}$ is retro-time elapsed to reach those configurations. In this case, we have that
\begin{equation}
\label{eq:adjointsolutionstributary}
\begin{split}
\lambda_{x} &= -\cos\left(\phi_d - \nu_p^*\tau \right),\:\:
\lambda_{y} = \sin\left(\phi_d - \nu_p^*\tau \right),\\
\lambda_\theta &= \nu_e^* \left(-\sin(\phi_d -\theta_d) + \sin(\phi_d-\theta_d- \nu_e^*(\tau-\tau_{US}))\right),
\end{split}
\end{equation}
for $\tau \geq \tau_{US}$. Note that the evader applies a particular control at each side of the EUS, i.e., $\nu_e^*=-1$ or $\nu_e^*=1$.

The previous equations are valid as long as the players do not switch controls. Similarly to the primary surface, we found that the pursuer switches its control after some time.

In Section \ref{sec:simulation}, we have included two examples of the Universal Surface and its tributary trajectories. A description of the players' motions in the realistic space for both cases is presented.

\begin{figure}[t]
    \centering
    \includegraphics[scale=0.35]{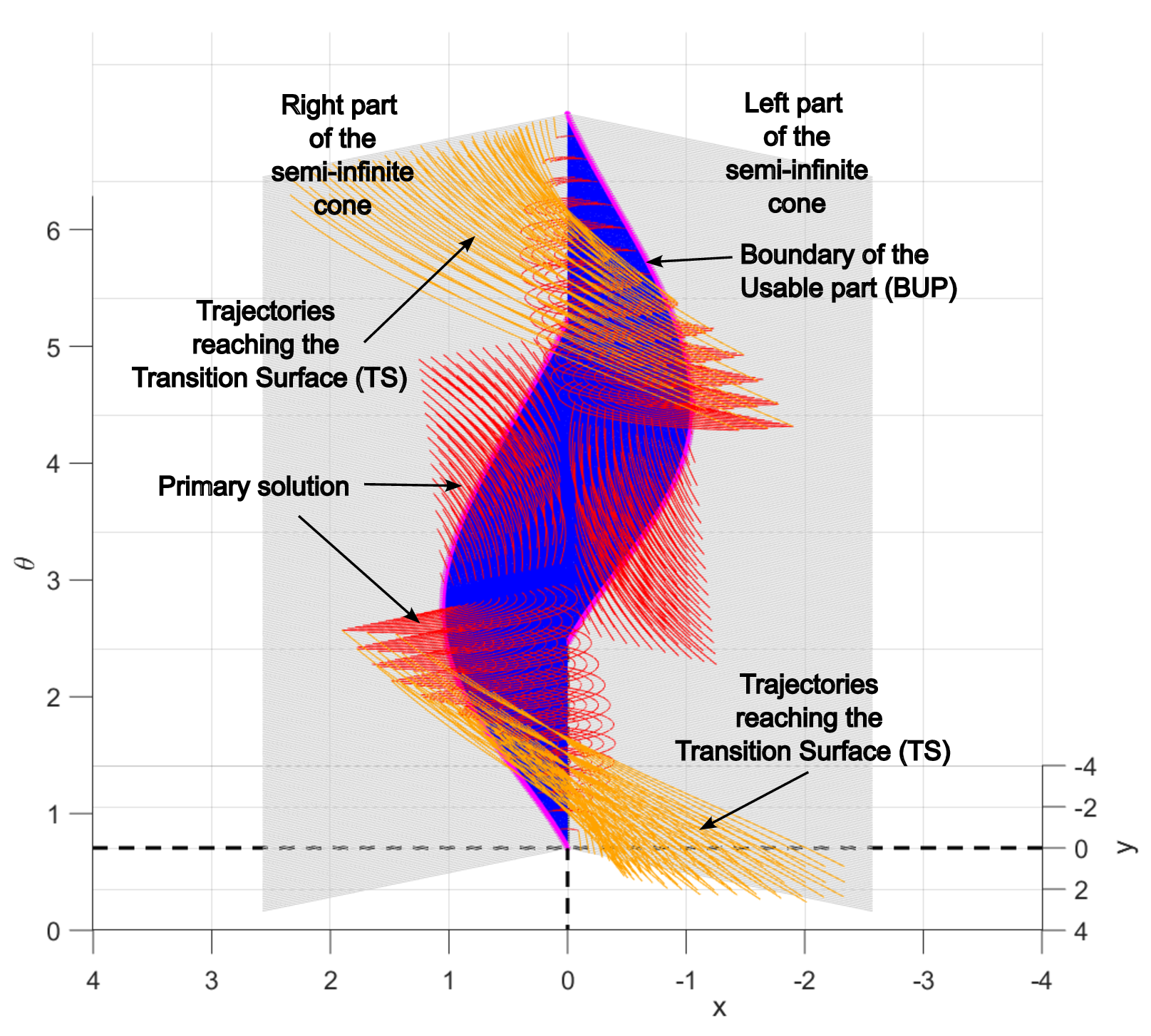}
    \caption{An example of the trajectories reaching the Transition Surface (orange curves) for $\phi_d=40^\circ$. The Transition surface corresponds to the points where the orange curves meet the primary solution (red curves).}
    \label{fig:transition_surface}
\end{figure}

\subsection{Transition Surface at the primary solution}

As mentioned before, we found that the pursuer switches 
control after some time when the system follows various trajectories of the primary solution (see Fig. \ref{fig:transition_surface}). We denote this time as $\tau_{s}$ and the configurations in the playing space where this change occurs belong to the Transition Surface (TS). Since transcendental equations describe the motion trajectories in the primary solution, we did not find an analytical expression for $\tau_s$. Thus, we employ numerical analysis to determine its value. When the system reaches $\tau_s$, we need to perform a new integration of the costate and motion equations taking as initial conditions the values of $\lambda_x$, $\lambda_y$, $\lambda_{\theta}$, $x$, $y$ and $\theta$ at $\tau_s$. In this case, the costate variables are given by the following expressions
\begin{equation}
\begin{split}
\lambda_x &= -\cos(\phi_d-\nu_{p_0}^*\tau_s-\nu_p^*(\tau-
\tau_s)), \\
\lambda_y &= \sin(\phi_d-\nu_{p_0}^*\tau_s-\nu_p^*(\tau-\tau_s)), \\
\lambda_{\theta} &= \nu_e^*(-\sin(\phi_d-\theta_d)+\sin(\phi_d-\theta_d-\nu_e^*\tau)),
\end{split}
\end{equation}
where $\nu_{po}^*$ denotes the pursuer's optimal control before the switch and $\nu_p^*$ is the pursuer's optimal control after the switch. For $\phi\in[0,\phi_d]$, we that $\nu_p^*$ switches from $-1$ to $1$, i.e., $\nu^*_{p_0}=-1$ and $\nu_p^*=1$ after the switch.

Integrating the motion equations, we get that
\begin{equation}
\label{eq:transition_primary}
\begin{split}
x &= -\nu^*_p+\nu^*_p\cos\left(\nu^*_p (\tau-\tau_s) \right)-\nu_e^*\cos\left(\theta_s-\nu^*_p (\tau-\tau_s)\right)\\
&+\nu_e^*\cos\left(\theta_s+\left(\nu^*_e - \nu^*_p\right)(\tau-\tau_s)\right)+r_s\sin\left(\phi_s- \nu^*_p(\tau-\tau_s)\right),\\
y &= \nu_p^*\sin\left(\nu^*_p(\tau-\tau_s)\right)+r_s\cos\left(\phi_s - \nu^*_p(\tau-\tau_s)\right)\\
&-2\nu_e^*\cos\left(\theta_s+\left(\frac{\nu^*_e}{2}-\nu^*_p\right)(\tau-\tau_s)\right)\sin\left(\frac{\nu^*_e}{2}(\tau-\tau_s)\right),\\
\theta &= \theta_s+\left(\nu^*_e-\nu^*_p \right)(\tau-\tau_s),
\end{split}
\end{equation}
where $(r_s,\phi_s,\theta_s)$ are the cylindrical coordinates of the system's state at time $\tau_s$ in the primary solution. Those expressions provide the trajectories emanating from the TS in retro-time. An instance of a trajectory reaching the Transition Surface and its continuation in the primary solution is presented in Section \ref{sec:simulation}. A description of players' motions in the realistic space is given.

\subsection{Transition Surface at the tributary trajectories of the Evader's Universal Surface}

\begin{figure}[t]
    \centering
    \includegraphics[scale=0.35]{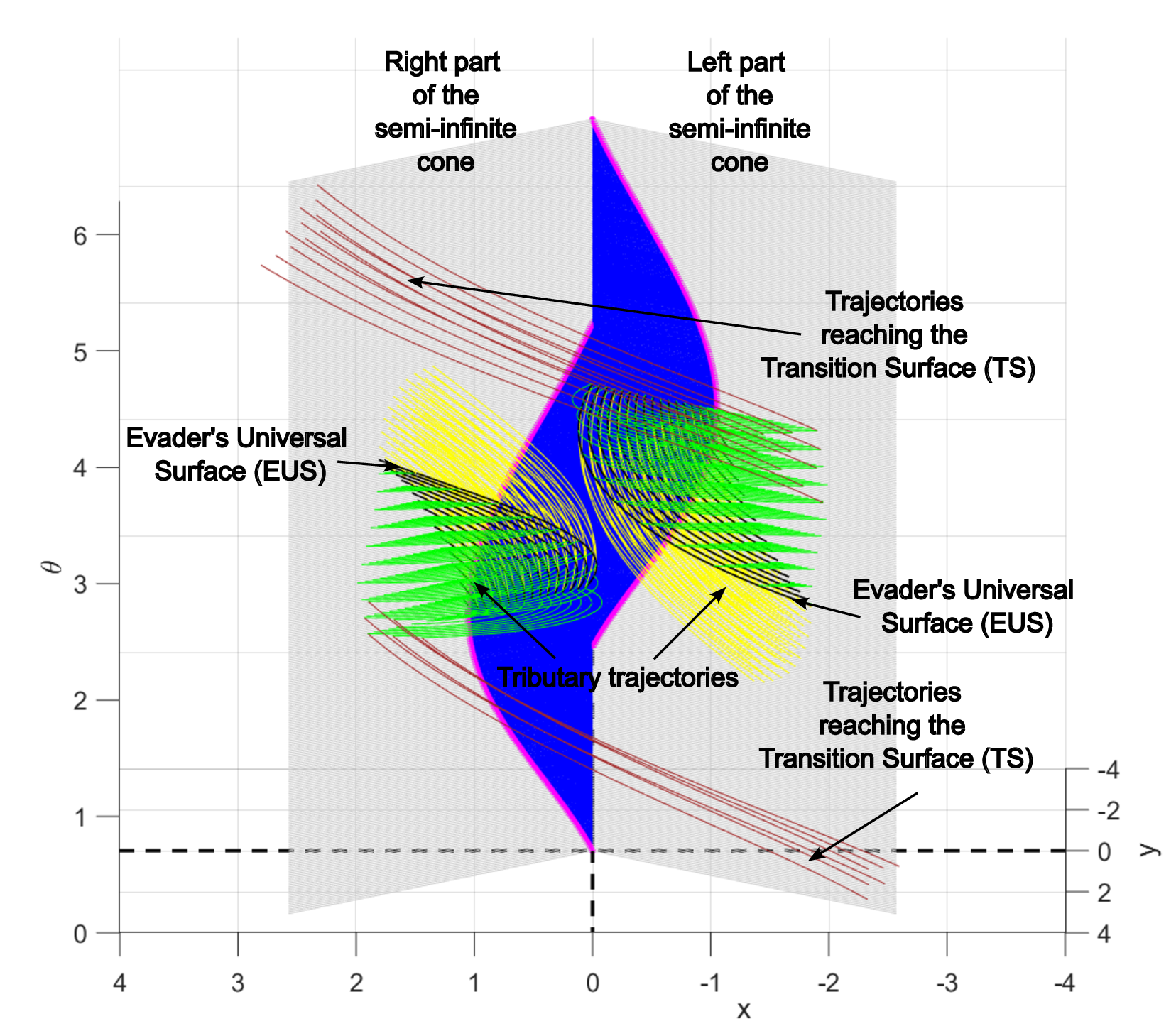}
    \caption{An example of the trajectories reaching the Transition Surface (brown curves) at the Evader's Universal Surface tributary trajectories for $\phi_d=40^\circ$. The Transition Surface corresponds to the points where the brown curves meet the tributary trajectories (yellow and green curves).}
    \label{fig:transition_surface_us}
\end{figure}

Similarly to the previous case, we found that for some tributary trajectories of the EUS, the pursuer switches control after some time $\tau_{s'}$ (see Fig. \ref{fig:transition_surface_us}). Thus, we need to perform a new integration of the motion and adjoint equations. Again, since the tributary trajectories are described by transcendental equations, we cannot find an analytical expression for $\tau_{s'}$. However, it can be computed numerically. Performing a new integration of the motion equations we have that
\begin{equation}
\label{eq:transition_tributary}
\begin{split}
x &= -\nu^*_p+\nu^*_p\cos\left(\nu^*_p (\tau-\tau_{s'}) \right)-\nu_e^*\cos\left(\theta_{s'}-\nu^*_p (\tau-\tau_{s'})\right)\\
&+\nu_e^*\cos\left(\theta_{s'}+\left(\nu^*_e - \nu^*_p\right)(\tau-\tau_{s'})\right)+r_{s'}\sin\left(\phi_{s'}- \nu^*_p(\tau-\tau_{s'})\right),\\
y &= \nu_p^*\sin\left(\nu^*_p(\tau-\tau_{s'})\right)+r_{s'}\cos\left(\phi_{s'} - \nu^*_p(\tau-\tau_{s'})\right)\\
&-2\nu_e^*\cos\left(\theta_{s'}+\left(\frac{\nu^*_e}{2}-\nu^*_p\right)(\tau-\tau_{s'})\right)\sin\left(\frac{\nu^*_e}{2}(\tau-\tau_{s'})\right),\\
\theta &= \theta_{s'}+\left(\nu^*_e-\nu^*_p \right)(\tau-\tau_{s'}),
\end{split}
\end{equation}
where $(r_{s'},\phi_{s'},\theta_{s'})$ are the cylindrical coordinates of the system's state at time $\tau_{s'}$ in the tributary trajectory. The costate variables are given by
\begin{equation}
\begin{split}
\lambda_x &= -\cos(\phi_d-\nu_{p_0'}^*\tau_s'-\nu_p^*(\tau-
\tau_s')), \\
\lambda_y &= \sin(\phi_d-\nu_{p_0'}^*\tau_s'-\nu_p^*(\tau-\tau_s')), \\
\lambda_\theta &= \nu_e^* \left(-\sin(\phi_d -\theta_d) + \sin(\phi_d-\theta_d- \nu_e^*(\tau-\tau_{US}))\right),
\end{split}
\end{equation}
where $\nu_{p_0'}^*$ denotes the pursuer's optimal control before the switch and $\nu_p^*$ is the pursuer's optimal control after the switch. For $\phi\in[0,\phi_d]$, we that $\nu_p^*$ switches from $-1$ to $1$, i.e., $\nu_{p_0'}=-1$ and $\nu_p^*=1$ after the switch.

\section{An attempt to solve the decision problem}
\label{sec:barrierfirst}

One of the main questions addressed when solving a pursuit-evasion game is determining the game's winner. In our problem, that means finding the region of initial configurations where the evader can escape surveillance and those where is impossible. In differential game theory, the curve separating those regions is known as {\em the barrier} \cite{ISAACS-99}. A similar approach to the one followed in the previous section is used to find the barrier. In this case, a retro-time integration of the costate and motion equations is performed, taking the configurations at the BUP as initial conditions. Recall that in our game, the BUP is composed of two sets, one corresponding to the right boundary of the detection region, denoted as RBUP and described by (\ref{eq:RBUP}), and the other corresponding to the left boundary of the detection region, denoted as LBUP and given by (\ref{eq:LBUP}).

\subsection{Primary solution}

Considering the configurations in the RBUP as initial conditions, i.e., $x_{RB}=r_{RB}\sin\phi_d$, $ y_{RB}=r_{RB}\cos\phi_d$ and $\theta=\theta_{RB}$, and performing the retro-time integration of the motion equations in (\ref{eq:retroreducedsystem}), we have that the barrier is described by the following equations
\begin{equation}
\label{eq:barrierprimary}
\begin{split}
x &= -\nu^*_p+{\nu^*_p}\cos\left(\nu^*_p \tau \right)-\nu_e^*\cos\left(\theta_{RB}-\nu^*_p \tau\right)+\nu_e^*\cos\left(\theta_{RB}+\left(\nu^*_e - \nu^*_p\right)\tau\right) \\
&+r_{RB}\sin\left(\phi_d- \nu^*_p\tau\right),\\
y &= \nu_p^*\sin\left(\nu^*_p\tau\right)+r_{RB}\cos\left(\phi_d- \nu^*_p\tau\right)-2\nu_e^*\cos\left(\theta_{RB}+\left(\frac{\nu^*_e}{2}-\nu^*_p\right)\tau\right)\sin\left(\frac{\nu^*_e}{2}\tau\right),\\
\theta &= \theta_{RB}+\left(\nu^*_e-\nu^*_p \right)\tau.
\end{split}
\end{equation}
where $\nu_p^*$ and $\nu_e^*$ are the players' optimal controls. In this case, we have that
\begin{equation}
\label{eq:adjointsolRB}
\begin{split}
\lambda_{x} &= -\cos\left(\phi_d - \nu_p^*\tau \right), \:\:
\lambda_{y} = \sin\left(\phi_d -  \nu_p^*\tau \right),\\
\lambda_\theta &= \nu_e^*\left(-\sin(\phi_d-\theta_{RB}) + \sin(\phi_d-\theta_{RB}- \nu_e^*\tau)\right).
\end{split}
\end{equation}

\subsection{Transition Surface}

Similar to the trajectories departing from the UP, substituting (\ref{eq:adjointsolRB}) and (\ref{eq:barrierprimary}) into (\ref{eq:pursuerctrls}), we found numerically that after following some of the barrier trajectories, the pursuer switches its control. We denote the time when the switch occurs as $\tau_{RBS}$. When the system reaches $\tau_{RBS}$, we need to perform a new integration of the costate and motion equations, taking as initial conditions the values of $\lambda_x$, $\lambda_y$, $\lambda_{\theta}$, $x$, $y$ and $\theta$ at $\tau_{RBS}$. In this case, the costate variables are given by the following expressions
\begin{equation}
\begin{split}
\lambda_x &= -\cos(\phi_d-\nu_{p_0}^*\tau_{RBS}-\nu_p^*(\tau-
\tau_{RBS})), \\
\lambda_y &= \sin(\phi_d-\nu_{p_0}^*\tau_{RBS}-\nu_p^*(\tau-\tau_{RBS})), \\
\lambda_{\theta} &= \nu_e^*(-\sin(\phi_d-\theta_{RB})+\sin(\phi_d-\theta_{RB}-\nu_e^*\tau)),
\end{split}
\end{equation}
where $\nu_{po}^*$ denotes the pursuer's optimal control before the switch and $\nu_p^*$ is the pursuer's optimal control after the switch. For $\phi\in[0,\phi_d]$, we that $\nu_p^*$ switches from $-1$ to $1$, i.e., $\nu^*_{p_0}=-1$ and $\nu_p^*=1$ after the switch. Integrating the motion equations, we get that
\begin{equation}
\label{eq:transition_barrier}
\begin{split}
x &= -\nu^*_p+\nu^*_p\cos\left(\nu^*_p (\tau-\tau_{RBS}) \right)-\nu_e^*\cos\left(\theta_{RBS}-\nu^*_p (\tau-\tau_{RBS})\right)\\
&+\nu_e^*\cos\left(\theta_{RBS}+\left(\nu^*_e - \nu^*_p\right)(\tau-\tau_{RBS})\right)+r_{RBS}\sin\left(\phi_{RBS}- \nu^*_p(\tau-\tau_{RBS})\right),\\
y &= \nu_p^*\sin\left(\nu^*_p(\tau-\tau_{RBS})\right)+r_{RBS}\cos\left(\phi_{RBS} - \nu^*_p(\tau-\tau_{RBS})\right)\\
&-2\nu_e^*\cos\left(\theta_{RBS}+\left(\frac{\nu^*_e}{2}-\nu^*_p\right)(\tau-\tau_{RBS})\right)\sin\left(\frac{\nu^*_e}{2}(\tau-\tau_{RBS})\right),\\
\theta &= \theta_{RBS}+\left(\nu^*_e-\nu^*_p \right)(\tau-\tau_{RBS}),
\end{split}
\end{equation}
where $(r_{RBS},\phi_{RBS},\theta_{RBS})$ are the cylindrical coordinates of the system's state at time $\tau_{RBS}$ in the primary solution. Similar results can be obtained for the configurations in the LBUP. In the following paragraphs, we present some interesting results regarding the process of constructing the barrier.

\subsection{Analysis of the BUPL}

As discussed before, a portion of the line formed by the intersection of the semi-infinite planes defining the detection region at $r=0$ belongs to the BUP, which we previously denoted as BUPL. From (\ref{eq:RBUP}), we have that for the RBUP, $r=0$ at $\theta=0$ and $\theta=\pi+2\phi_d$. Those configurations are part of the BUPL and are denoted as RBUPL to indicate they were obtained from the RBUP. In this section, we analyze if they produce a valid portion of the barrier. Note that a parallel examination needs to be done for the LBUP, since $r=0$ for $\theta=\pi-2\phi_d$ and $\theta=2\pi$.

On the RBUPL, we have that $\lambda_{\theta}=0$, $x=r\sin \phi_d=0$ and $y=r\cos \phi_d=0$ , thus $S=y\lambda_x -x\lambda_y + \lambda_\theta=0$. From (\ref{eq:pursuerctrls2}), $\nu_p^* = 0$, hence, we must check whether the pursuer continues using that control or switches it immediately. We can do that by analyzing the value of $\mathring S$ given by
\begin{equation}
\label{eq:dotS}    
\mathring S = \mathring y \lambda_x + y \mathring \lambda_y - \mathring x \lambda_y - x \mathring \lambda_y + \mathring \lambda_{\theta}.
\end{equation}
From (\ref{eq:retroreducedsystem}), on the RBUPL, 
\begin{equation}
\label{eq:retroRBUPL}
\begin{split}
\mathring{x} &= -\nu_p y - \sin \theta, \:\:
\mathring{y} = \nu_p x  + 1 - \cos \theta. \:\:
\end{split}
\end{equation}
Recall we have two cases, $\theta=0$ and $\theta=\pi+2\phi_d$. Substituting $\theta=0$ into (\ref{eq:adjointeq}) and (\ref{eq:retroRBUPL}), and the resulting expressions into (\ref{eq:dotS}), we obtain that $\mathring S =-\cos \phi_d.$ That indicates the pursuer must apply a control $\nu_p^*=-1$ immediately before the game's end. The pursuer's choice matches the evader's control, which is provided by (\ref{eq:lambdathfinal}), and also requests it to apply a control $\nu_e^*=-1$ immediately before the game's end. The previous result indicates that if both players consistently apply those controls exactly at the same time, the positions and orientations of both players permanently coincide, and thus, the evader is always in the boundary of the detection region. Note that in practice, this condition may be impossible to achieve.

Analogously, for $\theta=\pi+2\phi_d$, we found that $\mathring S = -\cos \phi_d$, which indicates again that the pursuer must apply a control $\nu_p^*=-1$ immediately before the game's end. The evader, on the other side, applies a control $\nu_e^*=1$, which is given by (\ref{eq:lambdathfinal}). Note that the previous results discard the existence of a Universal Surface in those cases. An analogous procedure can be applied to the LBUP, which also discards the existence of a Universal Surface for $r=0$ when $\theta=\pi-2\phi_d$ and $\theta=2\pi$.

\subsection{Direction of emanation of the barrier}

We found that the barrier's standard construction produces a surface that partially lies outside the playing space. To analyze this behavior, we employ the retro-time version of the motion equations in cylindrical coordinates,
\begin{equation}
\label{eq:polarsystemretro}
\begin{split}
\mathring r &= -\cos(\theta - \phi) + \cos \phi, \:\:
\mathring \phi = -\nu_p - \frac{\sin(\theta-\phi)+\sin \phi}{r}, \:\:
\mathring \theta = -\nu_p  + \nu_e.
\end{split}
\end{equation}

From (\ref{eq:BUPeq}), we know the barrier starts at those configurations in the boundary of the detection region where $\dot \phi = 0$, or $\mathring \phi = 0$, considering the retro-time version of the motion equations in (\ref{eq:polarsystemretro}). Thus, if $\ringring \phi < 0$ at the RBUP, the barrier trajectory will continue inside the playing space, and if $\ringring \phi > 0$, it will lay outside the detection region. From (\ref{eq:RBUP}), we have that $r=\sin(\theta-\phi_d)+\sin(\phi_d)$ at the RBUP, and recalling also that $\mathring \phi = 0$, we get that $\ringring \phi$ is given by
\begin{equation}
\ringring \phi = -\frac{(1+\mathring \theta)\cos(\theta_d-\phi_d) - \cos \phi_d}{r}.
\end{equation}
Now we proceed to determine under which conditions $\ringring \phi$ is positive or negative. From (\ref{eq:lambdathfinal}), we found that $\nu^*_e=-1$ for $\theta\in[0,\phi_d + \frac{\pi}{2})$, $\nu^*_e=0$ for $\theta=\phi_d+\frac{\pi}{2}$ and $\nu^*_e=1$ for $\theta\in(\phi_d+\pi/2, \pi + 2\phi_d]$. Recall that from (\ref{eq:RBUP}), the RBUP contains only configurations where $\theta\in[0,\pi+2\phi_d]$. Also, from (\ref{eq:pursuerctrls}), we have that $\nu^*_p=-1$ on the right boundary of the detection region. Substituting the players' controls described above into the expression for $\mathring \theta$ in (\ref{eq:polarsystemretro}) produces three cases. 1) If $\nu^*_e=-1$ then $\mathring \theta = 0$, therefore, the sign of $\ringring \phi$ is determined solely by the expression $-\cos(\theta-\phi_d)+\cos \phi_d$. Note that $-\cos(\theta-\phi_d)+\cos \phi_d=0$, for $\theta=2\phi_d$. In this case, we found that if $\theta \in (0, 2\phi_d)$, then $\ringring \phi < 0$, the barrier will continue inside the detection region. For $\theta \in [2\phi_d,\phi_d + \frac{\pi}{2})$, the barrier will continue outside the detection region and, therefore, is invalid. 2) If $\nu^*_e=0$ then $\mathring \theta = 1$, thus the sign of $\ringring \phi$ is determined by the expression $-2\cos(\theta-\phi_d)+\cos \phi_d$. Remember that, in this case, $\theta=\phi_d+\frac{\pi}{2}$, hence, substituting $\theta$ into the previous expression results in $\ringring \phi > 0$, which leads to a barrier that will continue outside the detection region. This case actually corresponds to a trajectory of the Universal Surface, described in the previous section, that immediately leaves the detection region. 3) If $\nu^*_3=1$ then $\ringring \theta = 3$, thus the sign of $\ringring \phi$ is determined by $-3\cos(\theta-\phi_d)+\cos \phi_d$. For this case, we found that $\ringring \phi > 0$ for $\theta\in(\phi_d + \frac{\pi}{2},\pi+2\phi_d)$, which produces a barrier that continues outside the detection region.

\begin{figure}[t]
    \centering
    \includegraphics[scale=0.37]{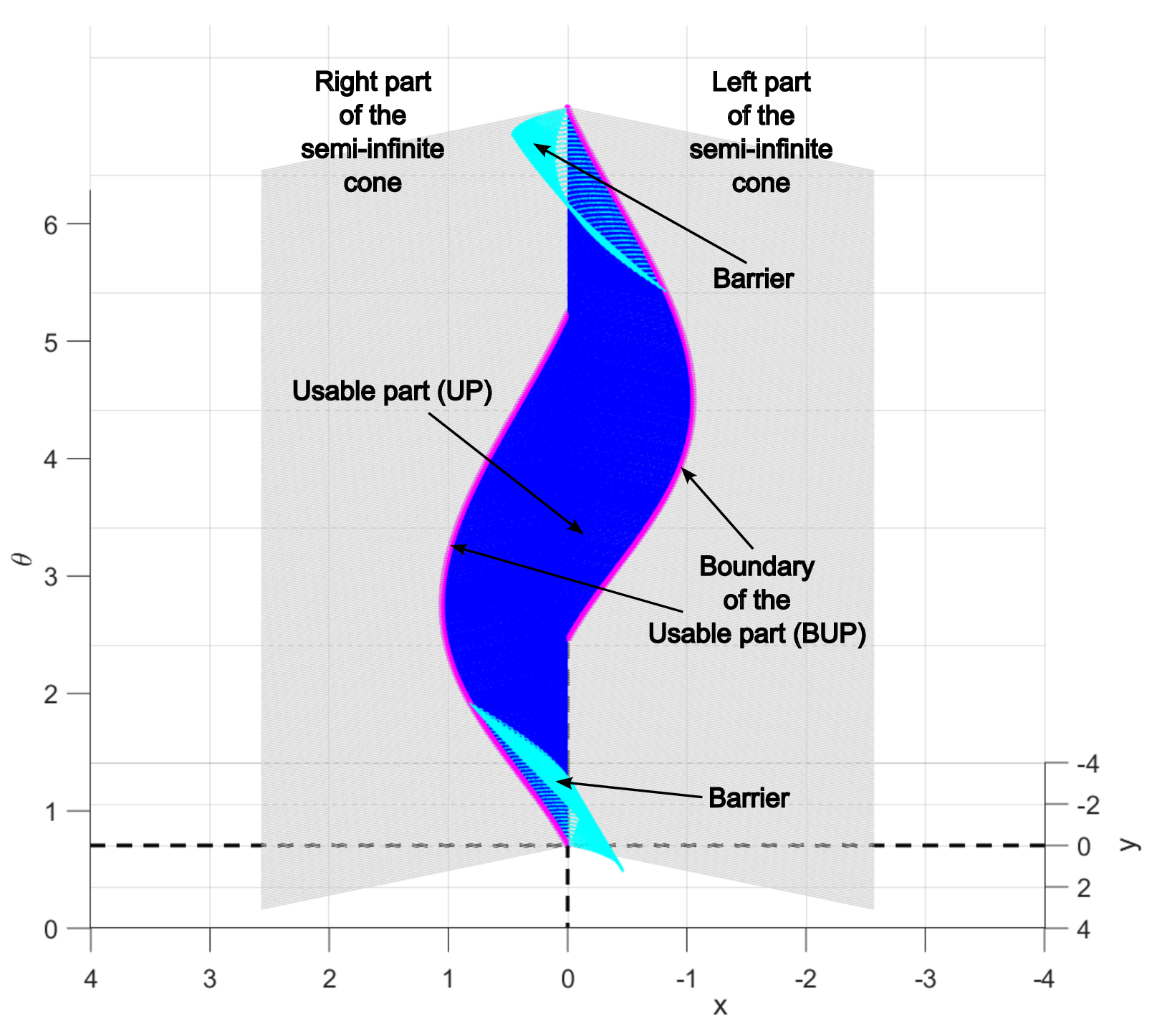}
    \caption{Representation of the barrier (cyan curves) for $\phi_d=40^\circ$. We can observe that some configurations at the BUP do not have a trajectory reaching it. The reason behind this is that the retro-integration of the motion equations starting from those configurations produces trajectories outside the semi-infinite cone, which are invalid.}
    \label{fig:barrier}
\end{figure}

Fig. \ref{fig:barrier} shows a representation of the barrier. In that figure, we can observe that only a subset of the configurations belonging to the BUP has a barrier trajectory that goes into the playing space, and those trajectories fail to define a closed region. This suggests that the evader can escape from all initial positions in the playing space or that the playing space is bounded by other barrier surfaces that also emerge from constructing additional singular surfaces. Unfortunately, in this paper, we cannot formally conclude which of the previous two cases occurs. A detailed analysis of the behavior of the barrier's trajectories outside the detection region maybe be useful to determine if those trajectories penetrate the detection region further in time and provide insight into missing parts of the barrier's trajectories. We left that study as future work.

\begin{figure}[t]
    \centering
    \includegraphics[scale=0.4]{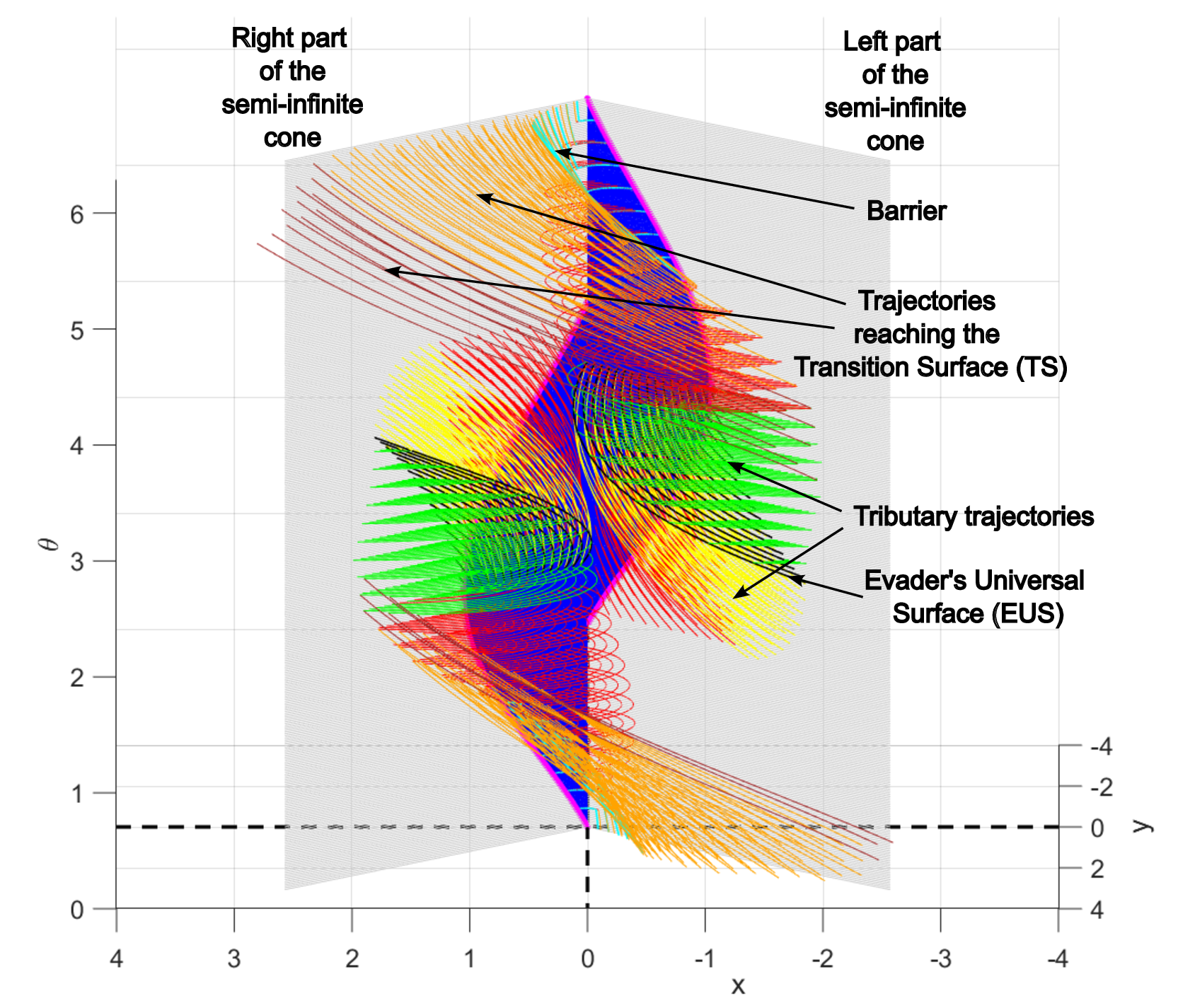}
    \caption{An overview of the trajectories found in this work near the game's end for $\phi_d=40^\circ$.}
    \label{fig:all_trajectories}
\end{figure}

\section{Discussion of the motion strategies near the game's end}

\label{sec:summary}

Fig. \ref{fig:all_trajectories} presents the set of motion strategies and their corresponding trajectories near the game's end. We can observe that the current trajectories are not enough to cover the entire playing space. This behavior suggests that additional singular surfaces must be found in the current problem. However, as pointed out before, finding them is an intricate task that may require a lot of algebraic manipulations and presumably numerical analysis. Recall that all players' trajectories presented in this work were obtained analytically, and they are represented by transcendental equations. From Fig. \ref{fig:all_trajectories}, we can notice that the tributary trajectories of the Evader's Universal Surface join smoothly with the trajectories of the primary solution. The same behavior can be observed with the trajectories reaching the Transition Surface. That indicates that the solutions seam those regions.

\section{Simulations}
\label{sec:simulation}

This section presents five numerical simulations to illustrate the players' motion strategies. The first two simulations exhibit two examples of the trajectories appearing in the primary solution. The third simulation displays a tributary trajectory of the Transition Surface followed by a trajectory in the primary solution. The fourth and fifth simulations correspond to two tributary trajectories reaching a particular point in the Universal Surface.

\begin{figure}[t]
    \centering
    \subfloat[Trajectory of the system in the reduced space. The red curve corresponds to the traveled trajectory belonging to the primary solution. \label{fig:primary1_reduced}]{
    \includegraphics[scale=0.28]{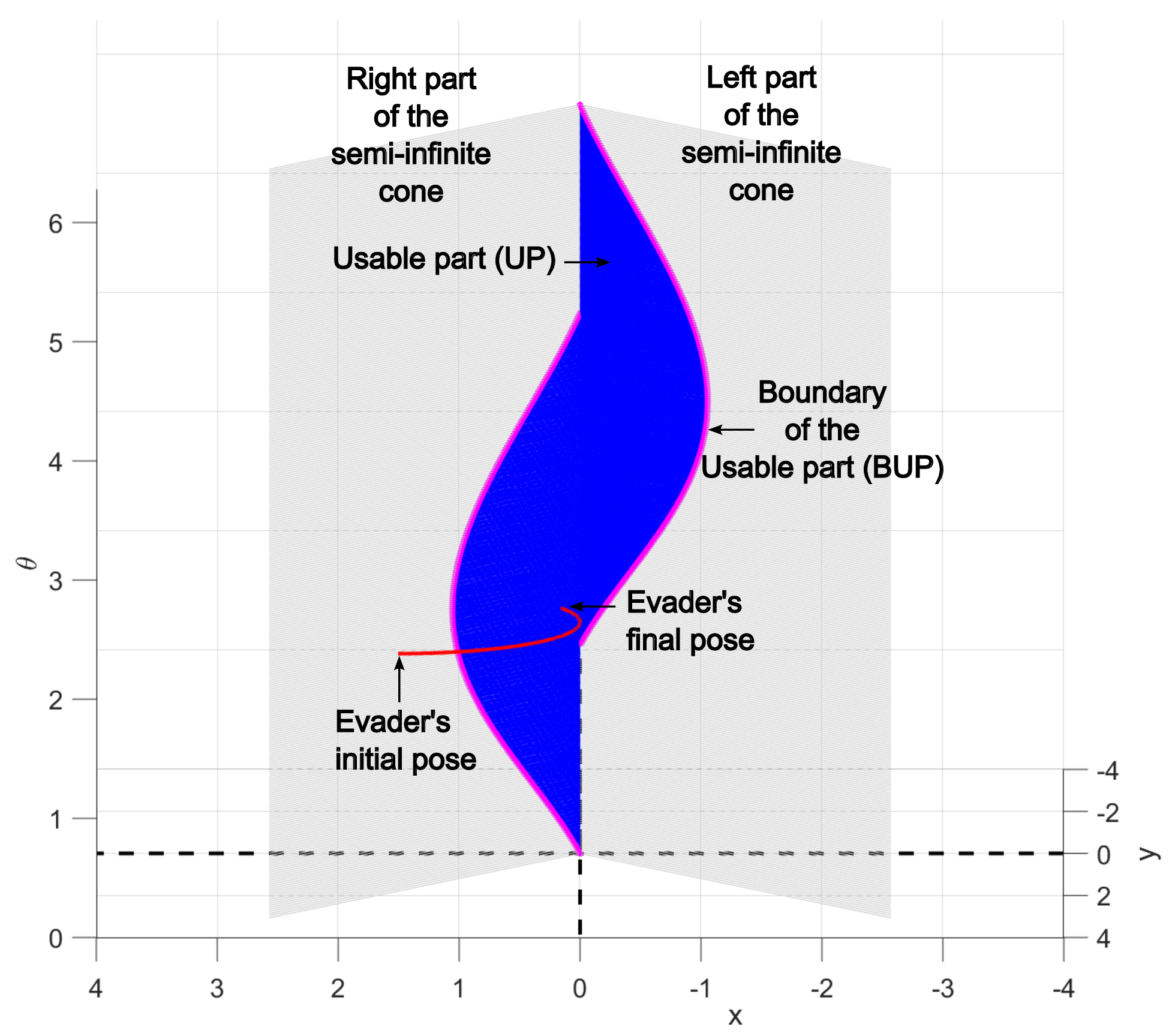}
    }
    \hfill
    \subfloat[Trajectory of the system in the realistic space. The blue curve indicates the pursuer's trajectory, and the red curve indicates the evader's trajectory. Arrows show the players' motion directions. \label{fig:primary1_realistic}]{
    \includegraphics[scale=0.42]{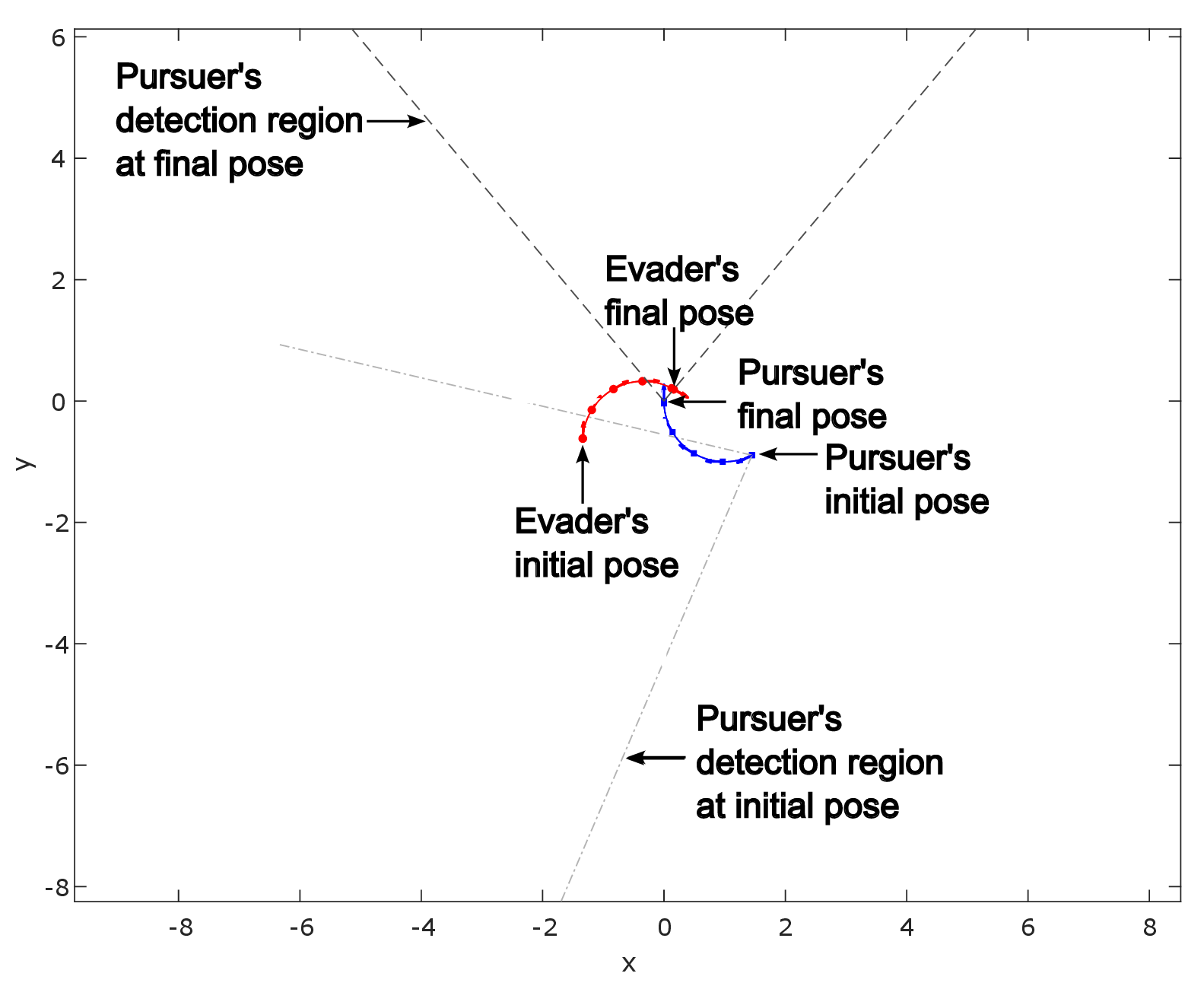}
    }
    \caption{A first example of a trajectory in the primary solution.}
    \label{fig:primary1}
\end{figure}

\begin{figure}[t]
    \centering
    \subfloat[Trajectory of the system in the reduced space. The red curve corresponds to the traveled trajectory belonging to the primary solution. \label{fig:primary2_reduced}]{
    \includegraphics[scale=0.28]{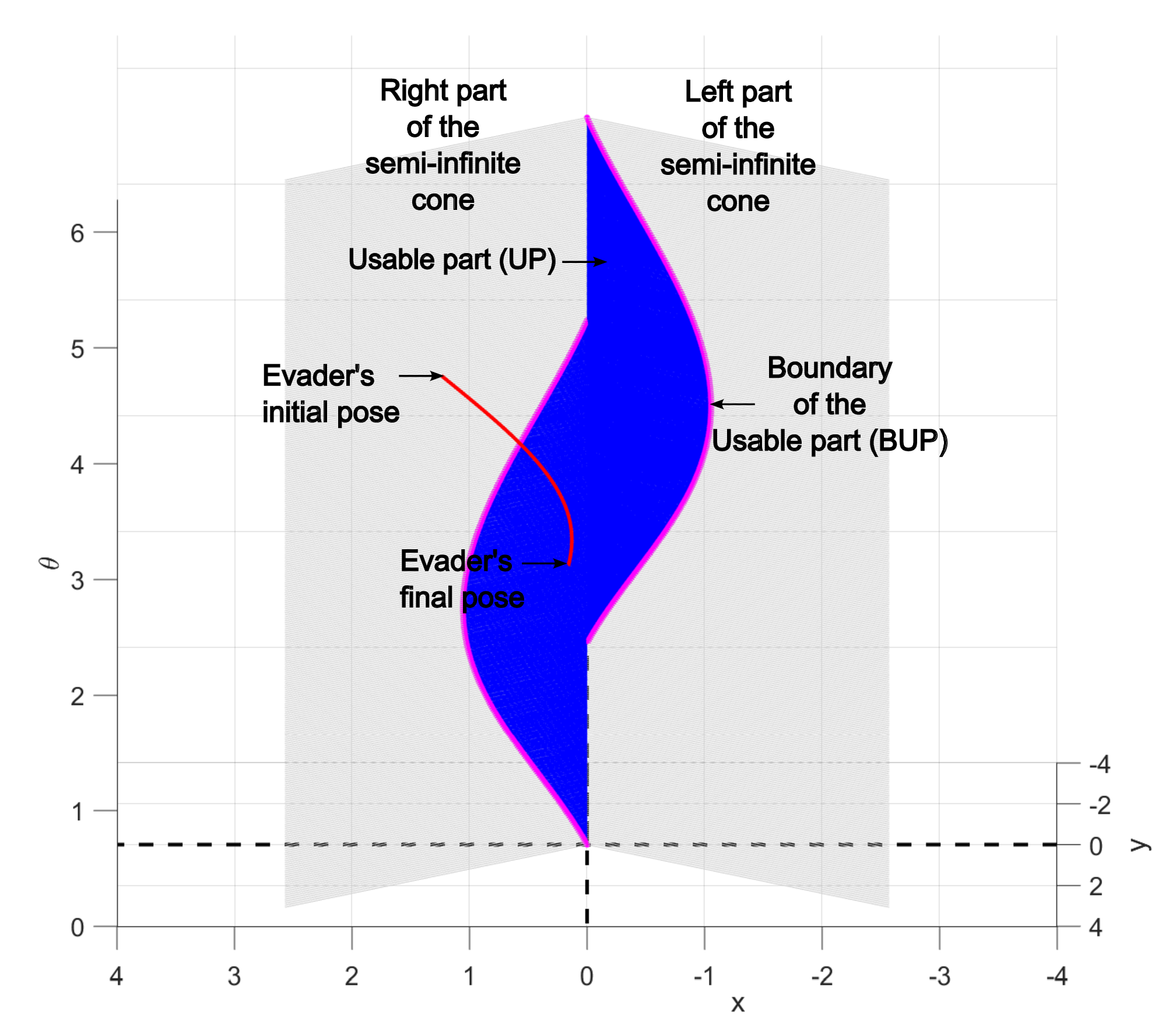}
    }
    \hfill
    \subfloat[Trajectory of the system in the realistic space. The blue curve indicates the pursuer's trajectory, and the red curve indicates the evader's trajectory. Arrows show the players' motion directions. \label{fig:primary2_realistic}]{
    \includegraphics[scale=0.42]{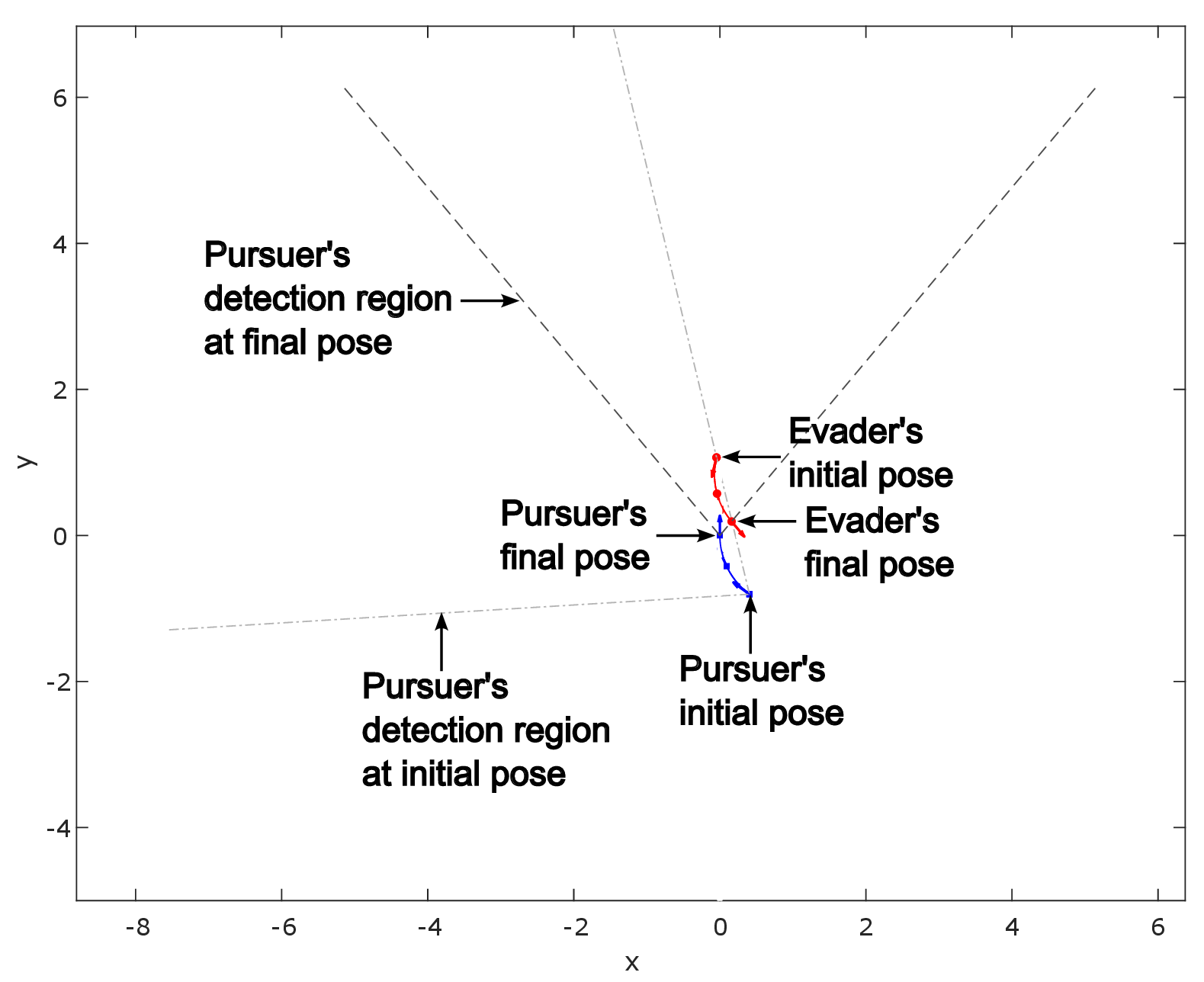}
    }
    \caption{A second example of a trajectory in the primary solution.}
    \label{fig:primary2}
\end{figure}

The parameters for the first simulation are $\phi_d=40^\circ$ and $\theta_d=120^\circ$. In this case, the evader starts close to the right boundary of the detection region (see Figs. \ref{fig:primary1_reduced} and \ref{fig:primary1_realistic}). In the reduced space, the trajectory followed by the system is represented by the red curve. In the realistic space (see Fig. \ref{fig:primary1_realistic}), the evader seeks to reach the right boundary of the detection region. The pursuer, on the contrary, pushes the right boundary away from the evader since it moves in that direction. However, despite the pursuer's efforts, the evader can reach the right boundary of the detection region and leave it. For the second simulation, we have that $\phi_d=40^\circ$ and $\theta_d=140^\circ$. In this case, the evader starts at the right boundary of the detection region (see Figs. \ref{fig:primary2_reduced} and \ref{fig:primary2_realistic}). In the realistic space (see Fig. \ref{fig:primary2_realistic}), we can observe that the evader, despite being located at the right boundary, has its heading pointing towards the interior of the detection region and since it cannot move backward, it cannot escape immediately. The evader aims to escape, moving in that direction and rotating counter-clockwise. On the contrary, the pursuer rotates clockwise and pushes away the boundary of the detection region. Eventually, the evader reaches the right boundary again but with a heading direction that allows it to escape.

\begin{figure*}[t]
\centering
\subfloat[Trajectory of the system in the reduced space. The orange curve corresponds to the traveled tributary trajectory of the Transition Surface, and the red curve corresponds to the traveled trajectory belonging to the primary solution. \label{fig:sim1_reduced}]{
\includegraphics[scale=0.4]{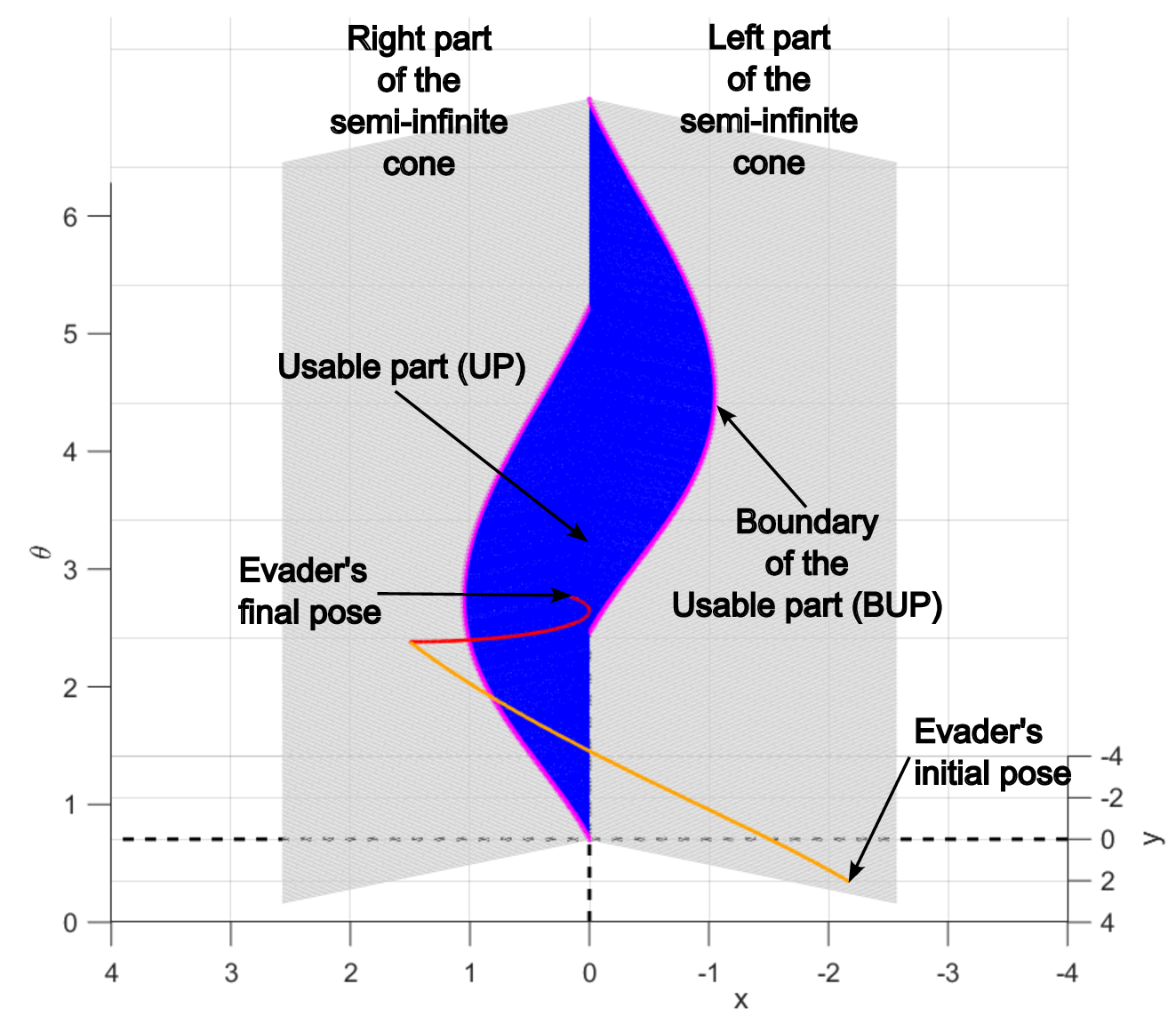}
}
\hfill
\subfloat[Trajectory of the system in the realistic space. The blue curve indicates the pursuer's trajectory, and the red curve indicates the evader's trajectory. Arrows show the players' motion directions. \label{fig:sim1_realistic}]{
\includegraphics[scale=0.6]{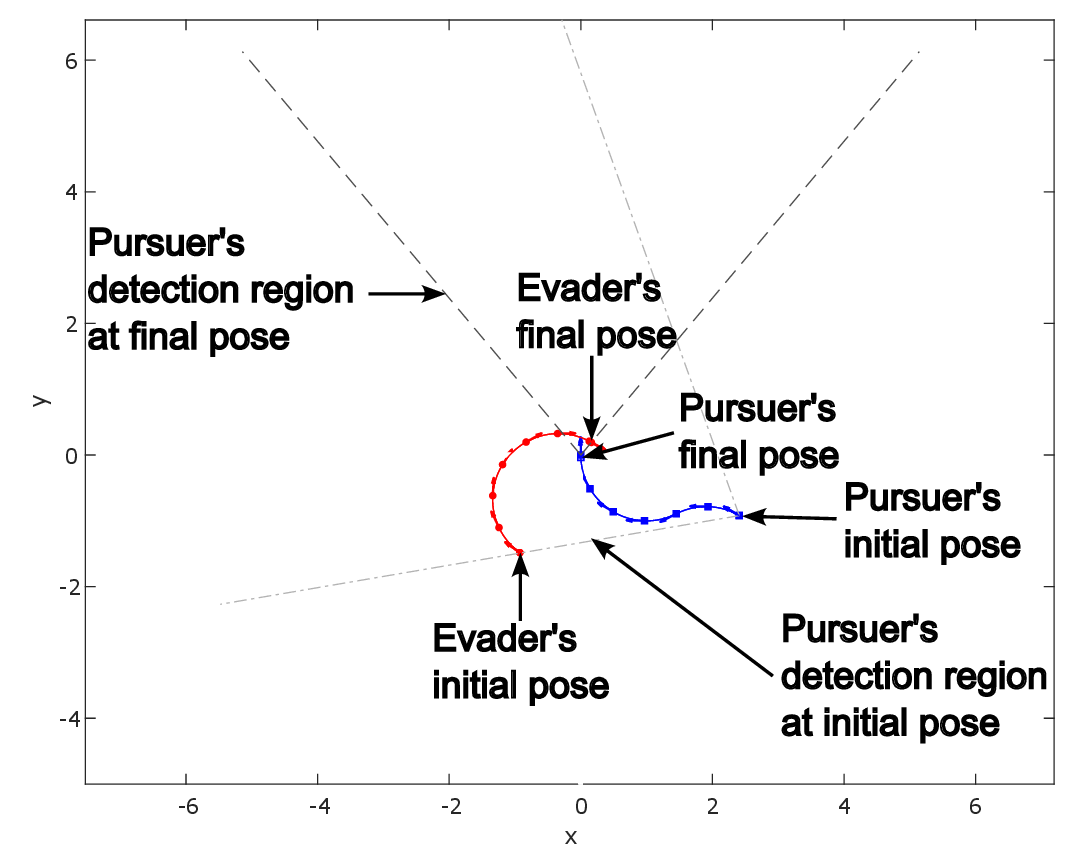}
}
\caption{An example of a tributary trajectory of the Transition Surface.} 
\end{figure*}

\begin{figure*}[h]
\centering
\subfloat[$t=0s$ \label{fig:sim1_seq1}]{
\includegraphics[scale=0.28]{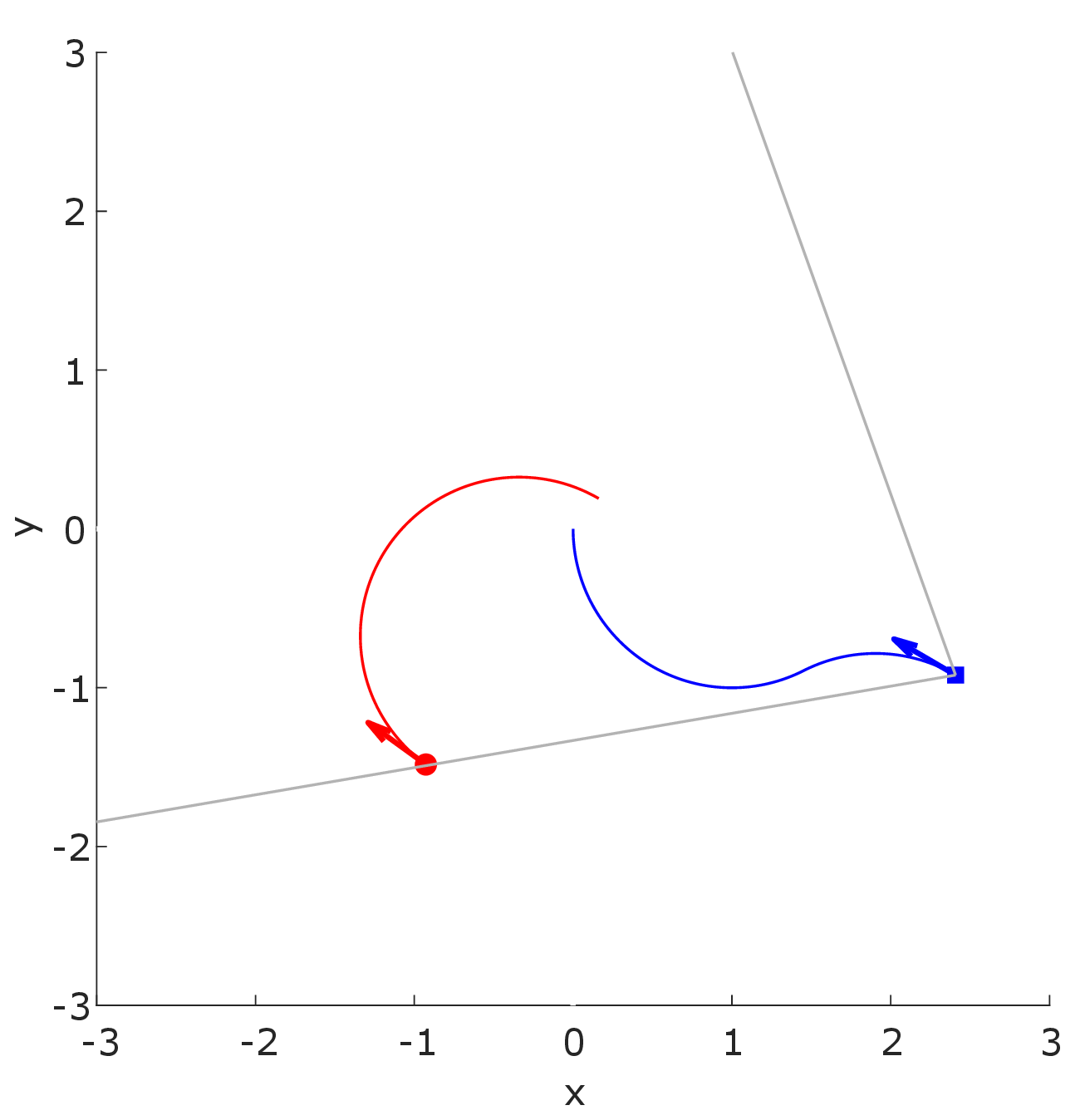}
}
\subfloat[$t=1s$ \label{fig:sim1_seq3}]{
\includegraphics[scale=0.28]{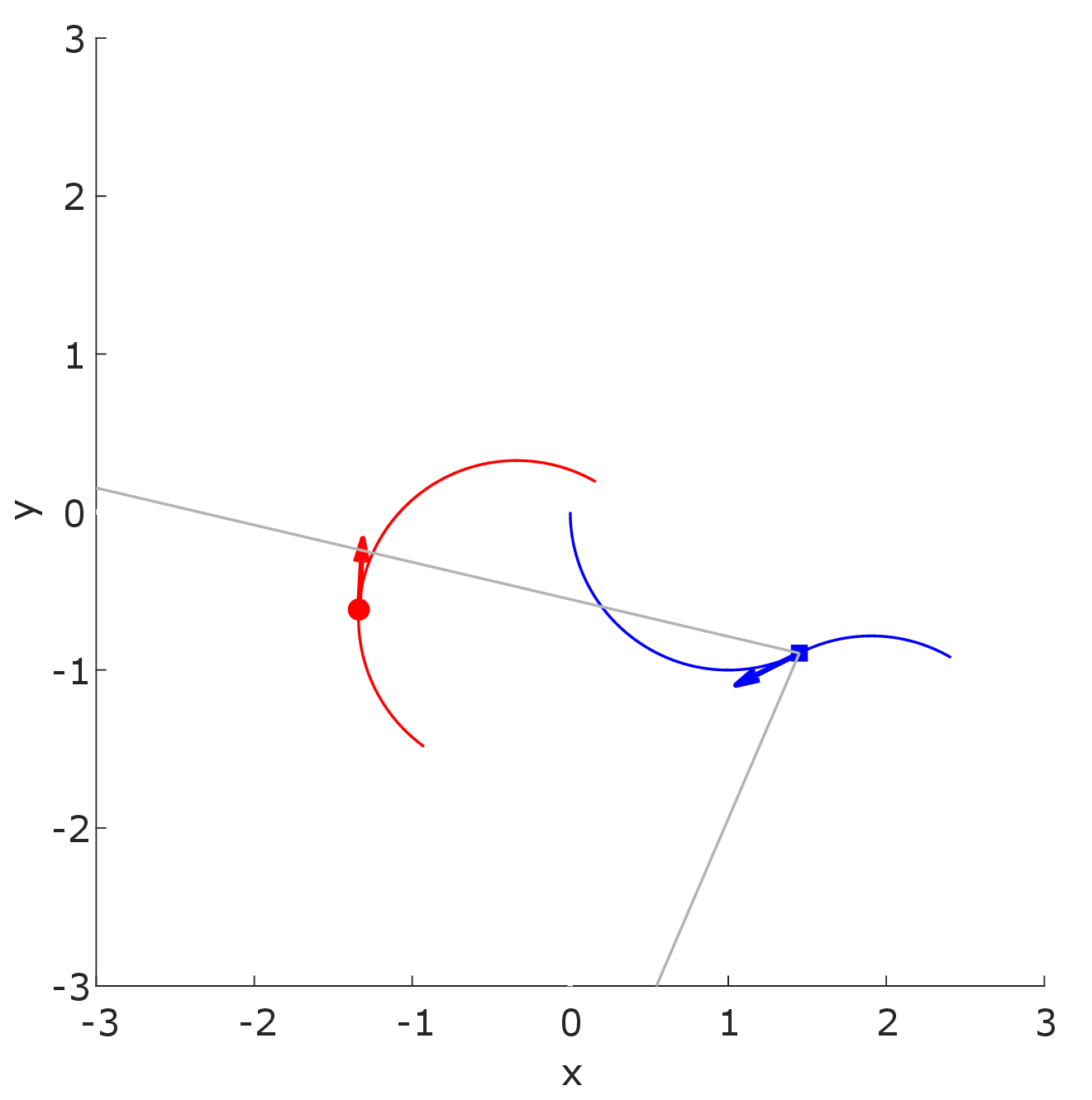}
}\\
\subfloat[$t=2s$ \label{fig:sim1_seq5}]{
\includegraphics[scale=0.28]{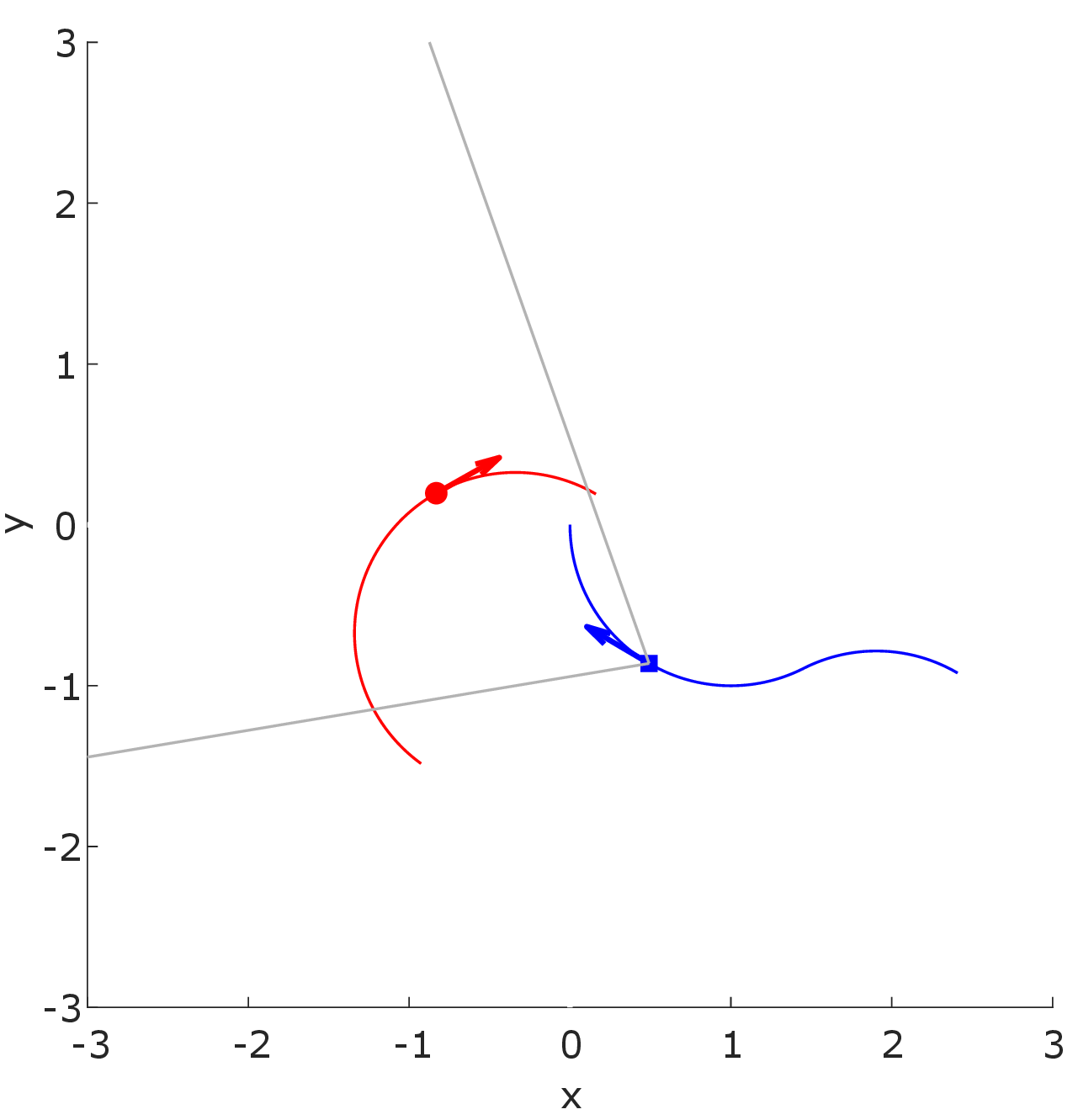}
}
\subfloat[$t=3.04s$ \label{fig:sim1_seq7}]{
\includegraphics[scale=0.28]{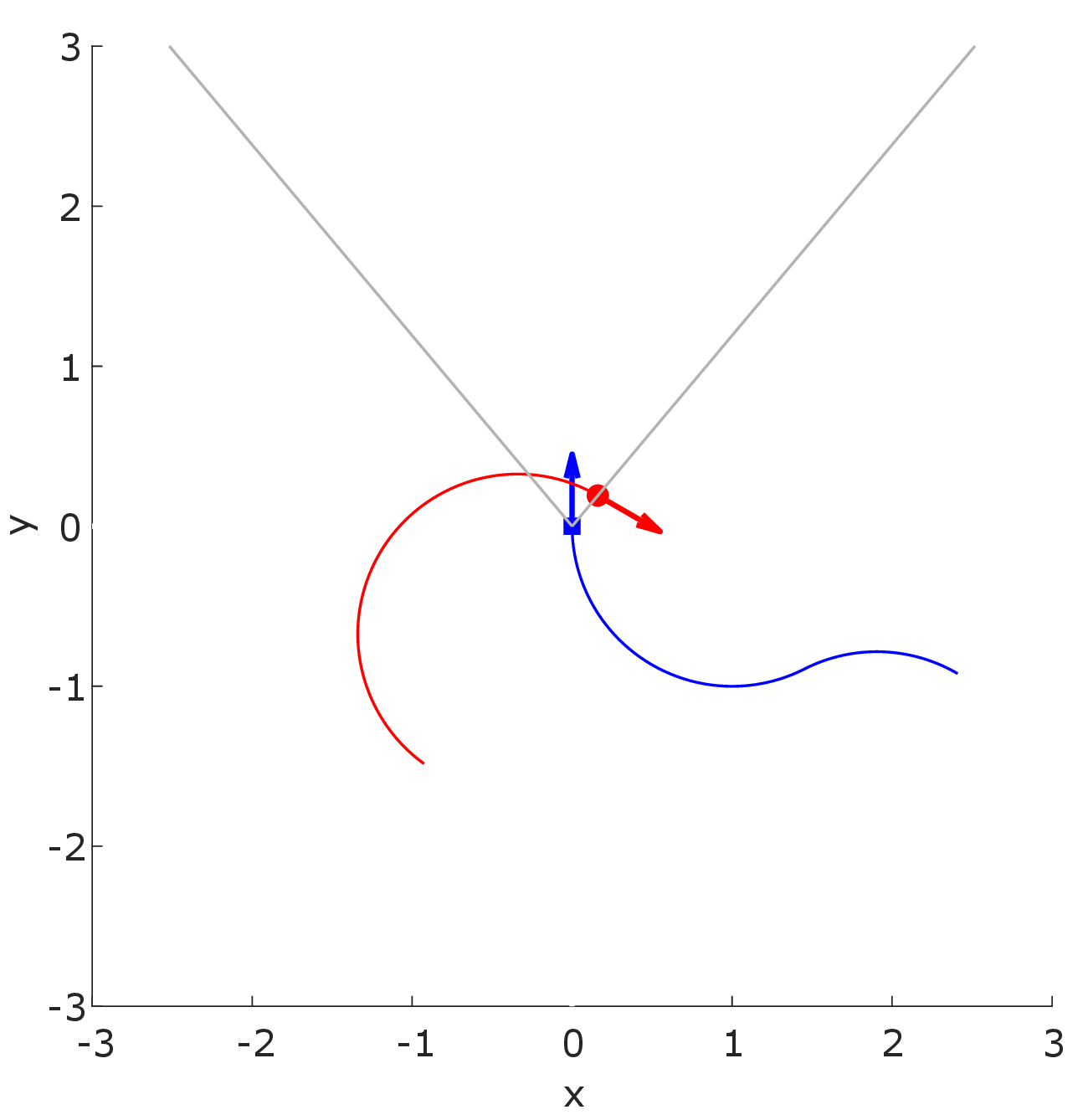}
}
\caption{Snapshots of the motion strategies of the players in the realistic space for the third simulation. The pursuer is represented by the blue dot, and the evader by the red one. The boundary of the FoV corresponds to the gray lines. The blue curve indicates the entire pursuer's trajectory, and the red one is the full evader's trajectory. The caption of each figure indicates the elapsed time of the simulation when it was taken. \label{fig:sim1snapshots}} 
\end{figure*}

The parameters for the third simulation are $\phi_d=40^\circ$ and $\theta_d=120^\circ$. In the example, the evader starts at the left boundary of the detection region (see Fig. \ref{fig:sim1_reduced}). In the reduced space, the system follows a trajectory (orange curve) that reaches the Transition Surface and continues to the terminal condition, traveling a trajectory (red curve) of the primary solution. In the realistic space (see Fig. \ref{fig:sim1_realistic}), the evader seeks to get closer to the pursuer and reach the right boundary of the detection region. Note that since the evader's initial orientation is pointing toward the interior of the left detection region, and it cannot move backward, it cannot escape immediately despite being located at the boundary. The pursuer takes advantage of this, first moving in a way that puts the evader in the center of the detection region and later pushing the right boundary away from the evader since it moves in that direction. However, despite the pursuer's efforts, the evader can reach the right boundary of the detection region and leave it. Fig. \ref{fig:sim1snapshots} shows several snapshots of the players' configurations in the realistic space taken during the first simulation.

\begin{figure*}[h]
\centering
\subfloat[Trajectory of the system in the reduced space. The green curve corresponds to the traveled tributary trajectory of the Universal Surface, and the black curve to the traveled portion of the Universal Surface. \label{fig:sim2_reduced}]{
\includegraphics[scale=0.4]{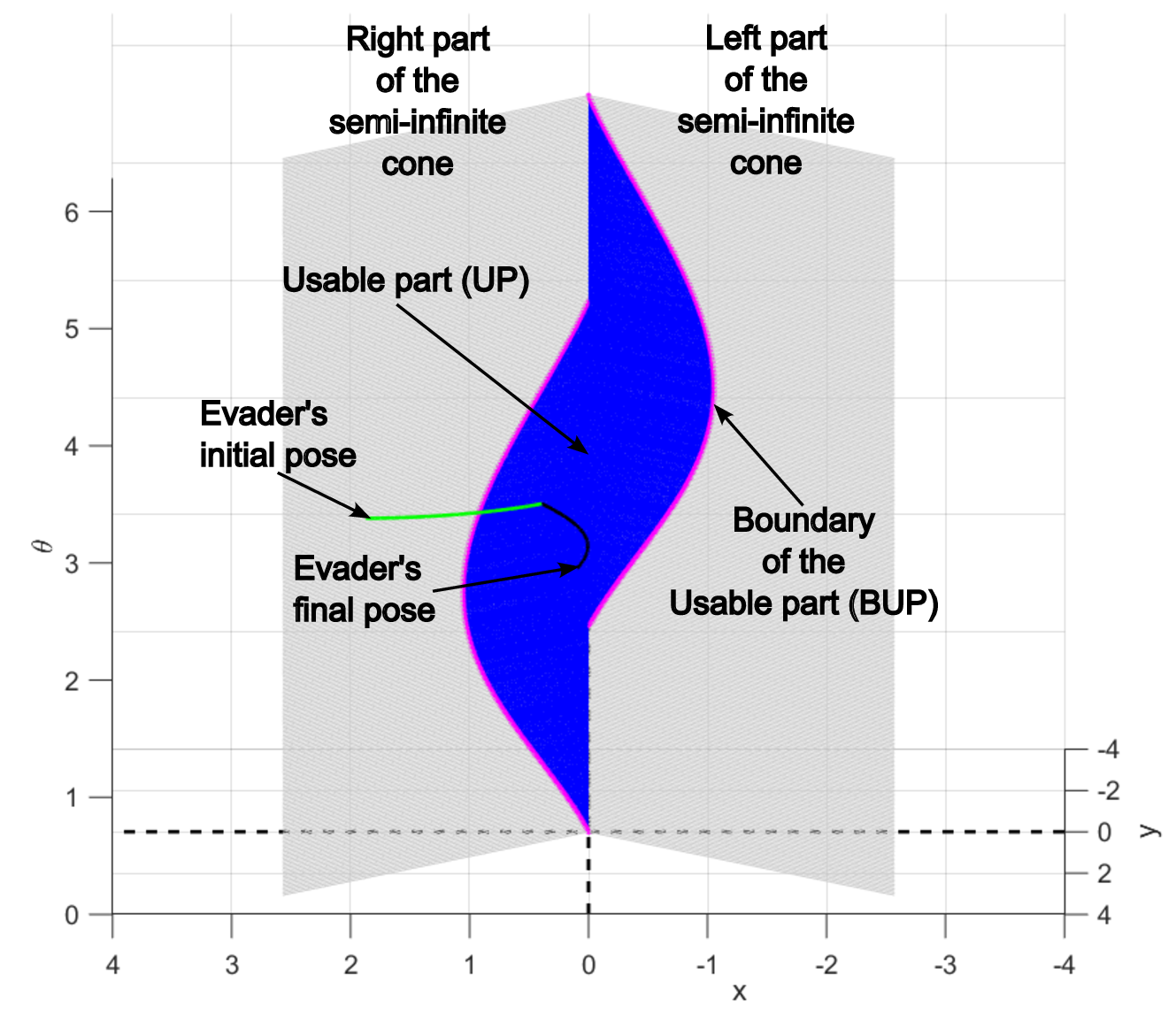}
}
\hfill
\subfloat[Trajectory of the system in the realistic space. The blue curve indicates the pursuer's trajectory and the red curve indicates the evader's trajectory. Arrows show the players' motion directions. \label{fig:sim2_realistic}]{
\includegraphics[scale=0.6]{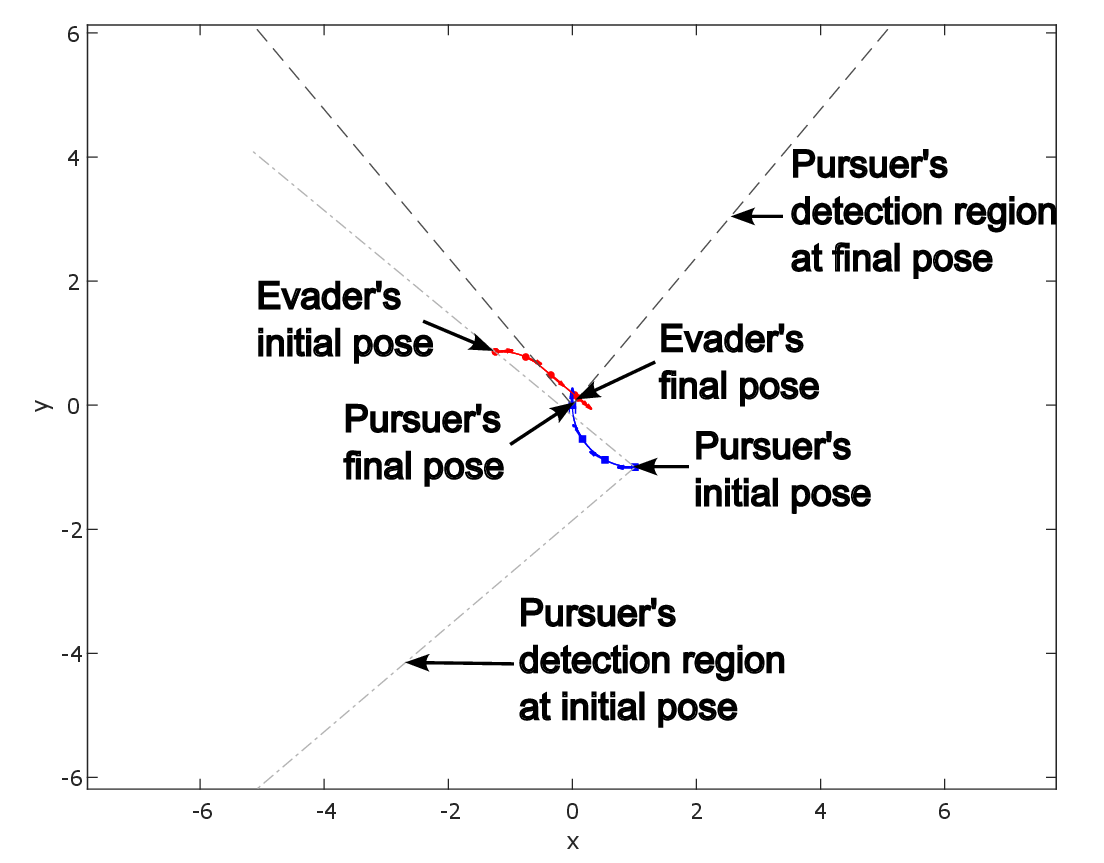}
}
\caption{A first example of a tributary trajectory of the Universal Surface.} 
\end{figure*}

\begin{figure*}[h]
\centering
\subfloat[$t=0s$ \label{fig:sim2_seq1}]{
\includegraphics[scale=0.28]{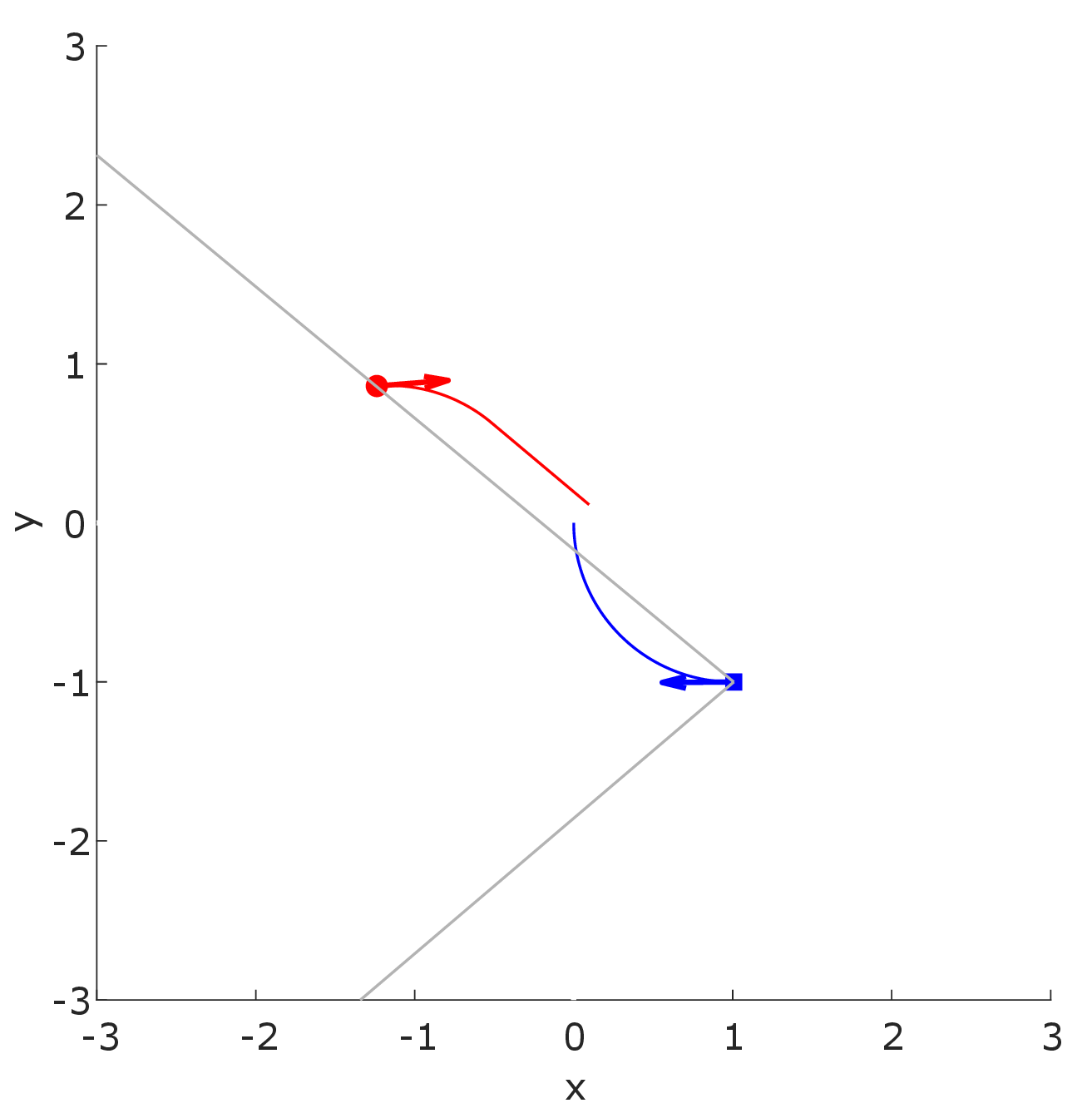}
}
\subfloat[$t=0.5s$ \label{fig:sim2_seq2}]{
\includegraphics[scale=0.28]{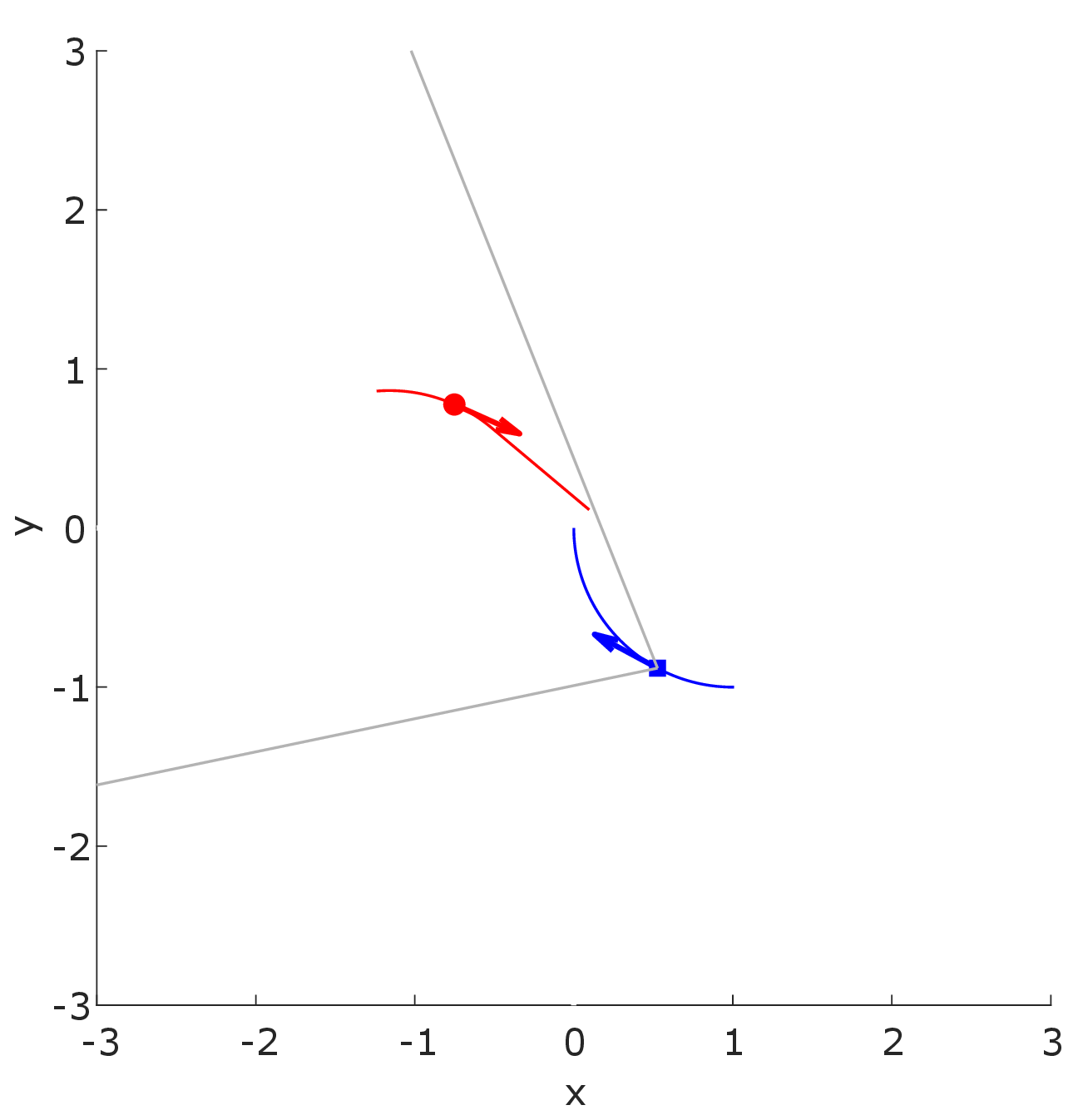}
}\\
\subfloat[$t=1s$ \label{fig:sim2_seq3}]{
\includegraphics[scale=0.28]{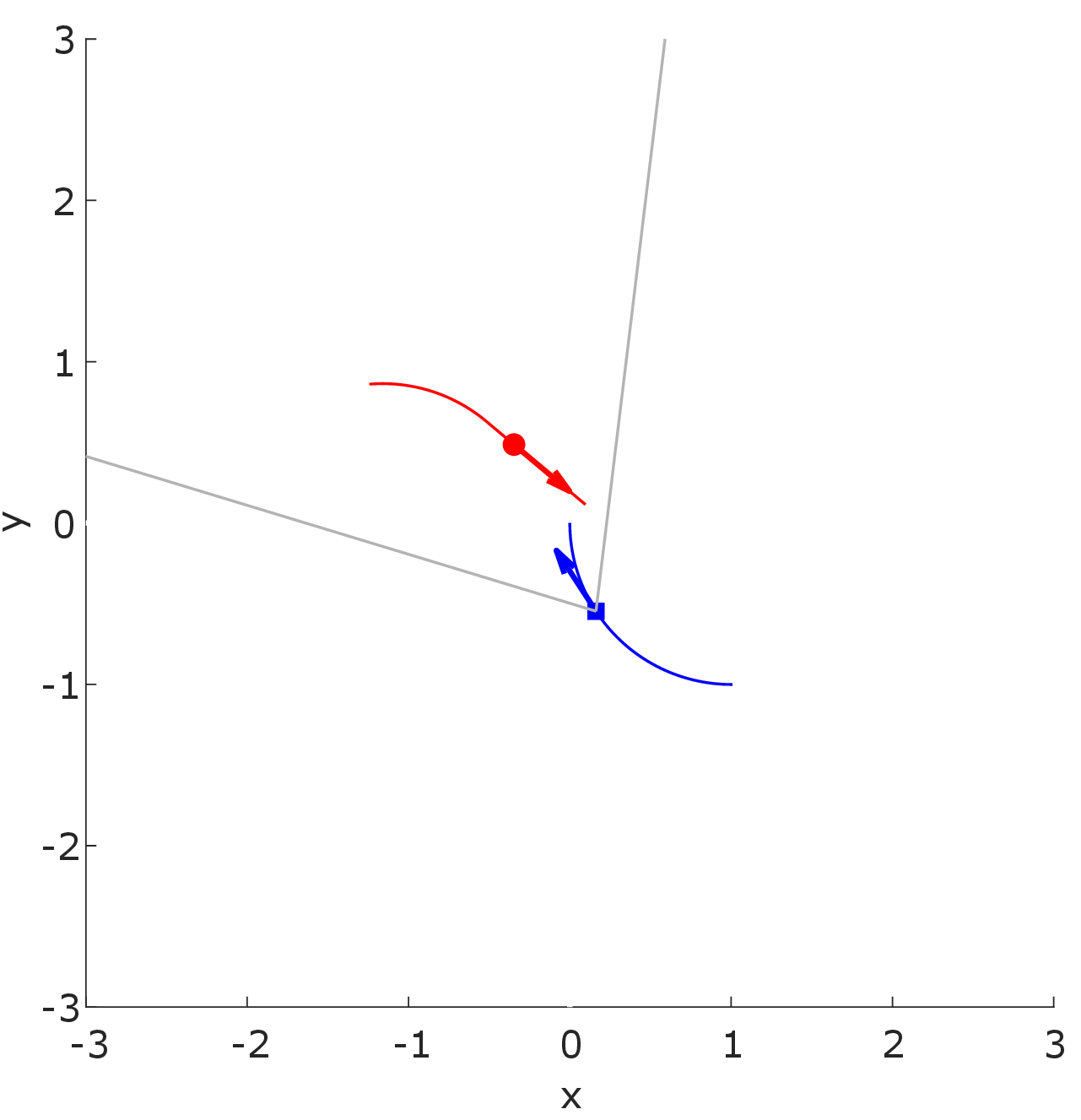}
}
\subfloat[$t=1.58s$ \label{fig:sim2_seq5}]{
\includegraphics[scale=0.28]{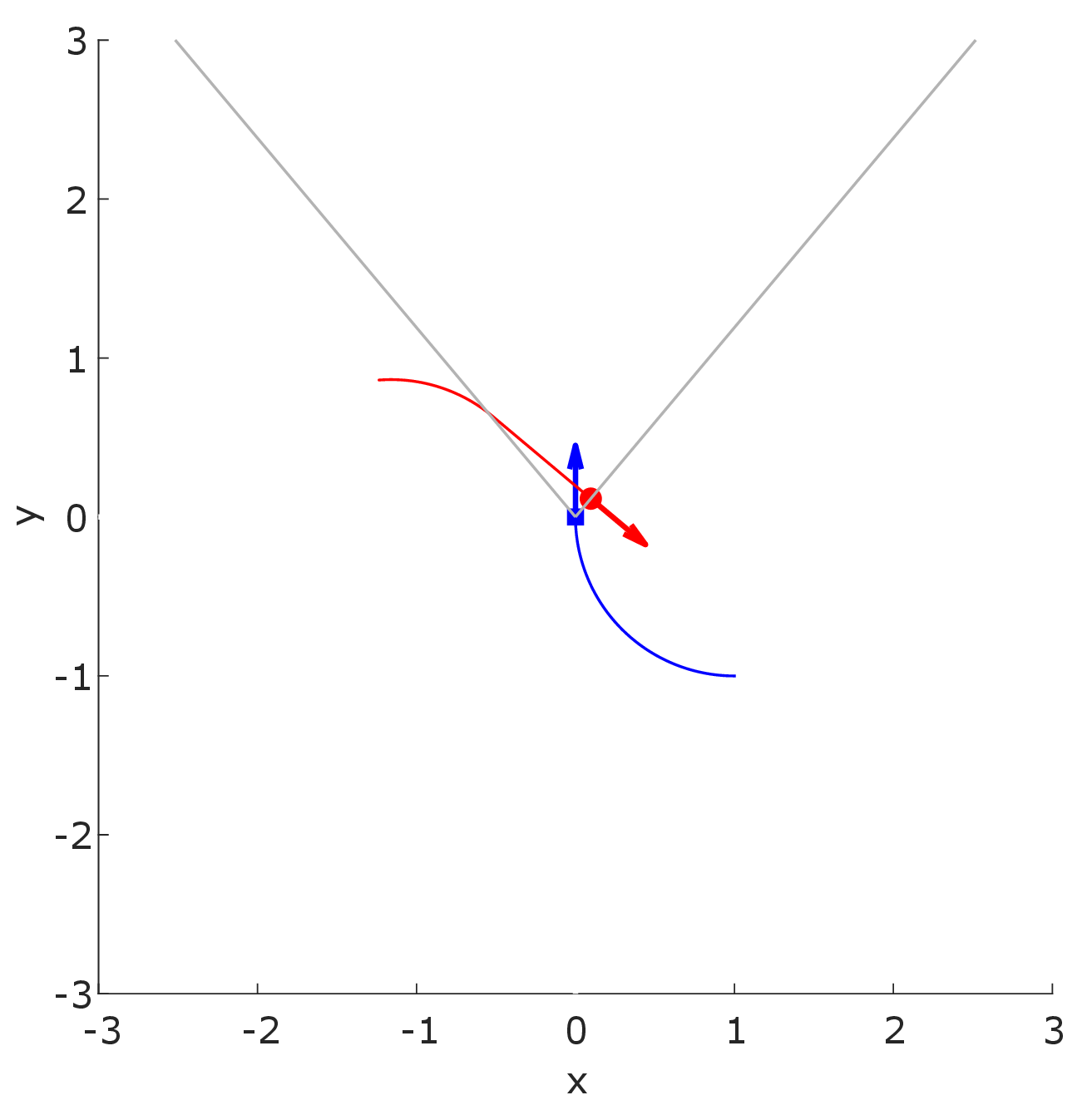}
}
\caption{Snapshots of the motion strategies of the players in the realistic space for the fourth simulation. The pursuer is represented by the blue dot, and the evader by the red one. The boundary of the FoV corresponds to the gray lines. The blue curve indicates the entire pursuer's trajectory, and the red one is the full evader's trajectory. The caption of each figure indicates the elapsed time of the simulation when it was taken. \label{fig:sim2snapshots}} 
\end{figure*}

We have $\phi_d=40^\circ$ and $\theta_d=\frac{\pi}{2}+\phi_d$ for the fourth simulation. In this example, the evader starts at the right boundary of the detection region (see Fig. \ref{fig:sim2_reduced}). In the reduced space, the system follows a tributary trajectory (green curve) of the Universal Surface and, after some time, continues to the terminal condition by traveling a portion of the Universal Surface (black curve). The evader's header points toward the detection region's exterior in the realistic space (see Fig. \ref{fig:sim2_realistic}). The evader seeks to escape by moving in that direction and rotating clockwise, but the pursuer counteracts the evader's motion also by rotating clockwise, pushing away the right boundary of the detection region. This pursuer's motion strategy momentarily avoids the evader's escape but allows it to get closer to the pursuer. After some time, the evader starts moving following a straight-line trajectory leading again to the right boundary of the detection region; this time, however, it can escape. Fig. \ref{fig:sim2snapshots} shows several snapshots of the players' configurations in the realistic space taken during the second simulation.

\begin{figure*}[h]
\centering
\subfloat[Trajectory of the system in the reduced space. The yellow curve corresponds to the traveled tributary trajectory of the Universal Surface, and the black curve to the traveled portion of the Universal Surface. \label{fig:sim3_reduced}]{
\includegraphics[scale=0.4]{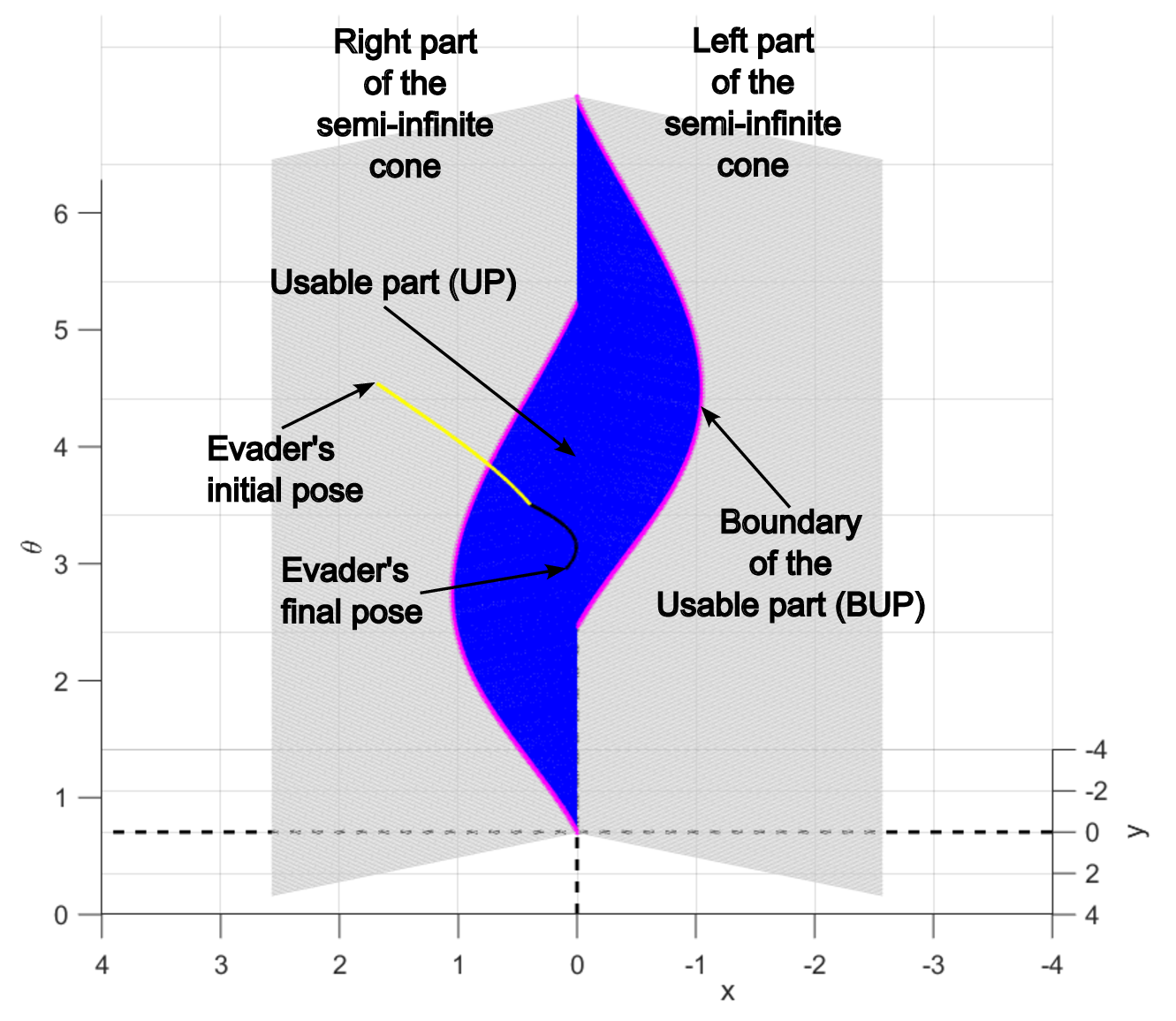}
}
\hfill
\subfloat[Trajectory of the system in the realistic space. The blue curve indicates the pursuer's trajectory, and the red curve indicates the evader's trajectory. Arrows show the players' motion directions. \label{fig:sim3_realistic}]{
\includegraphics[scale=0.6]{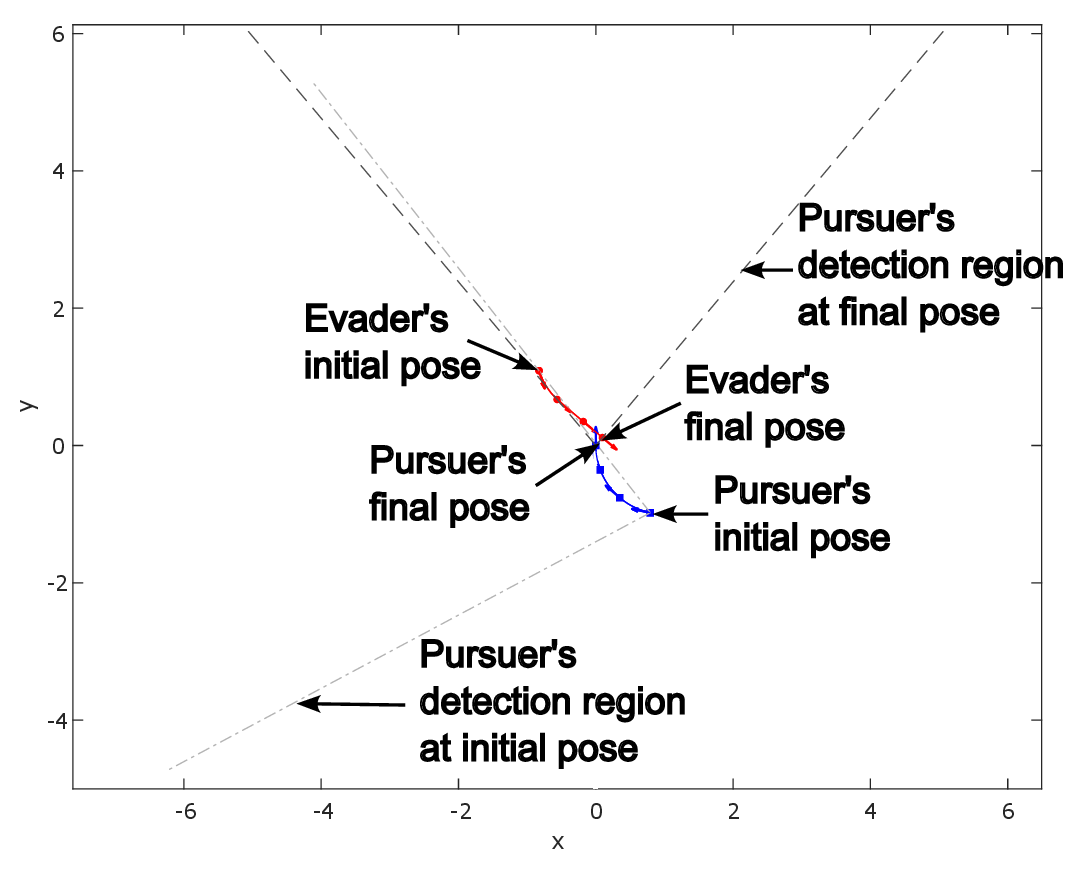}
}
\caption{A second example of a tributary trajectory of the Universal Surface.} 
\end{figure*}

\begin{figure*}[h]
\centering
\subfloat[$t=0s$ \label{fig:sim3_seq1}]{
\includegraphics[scale=0.28]{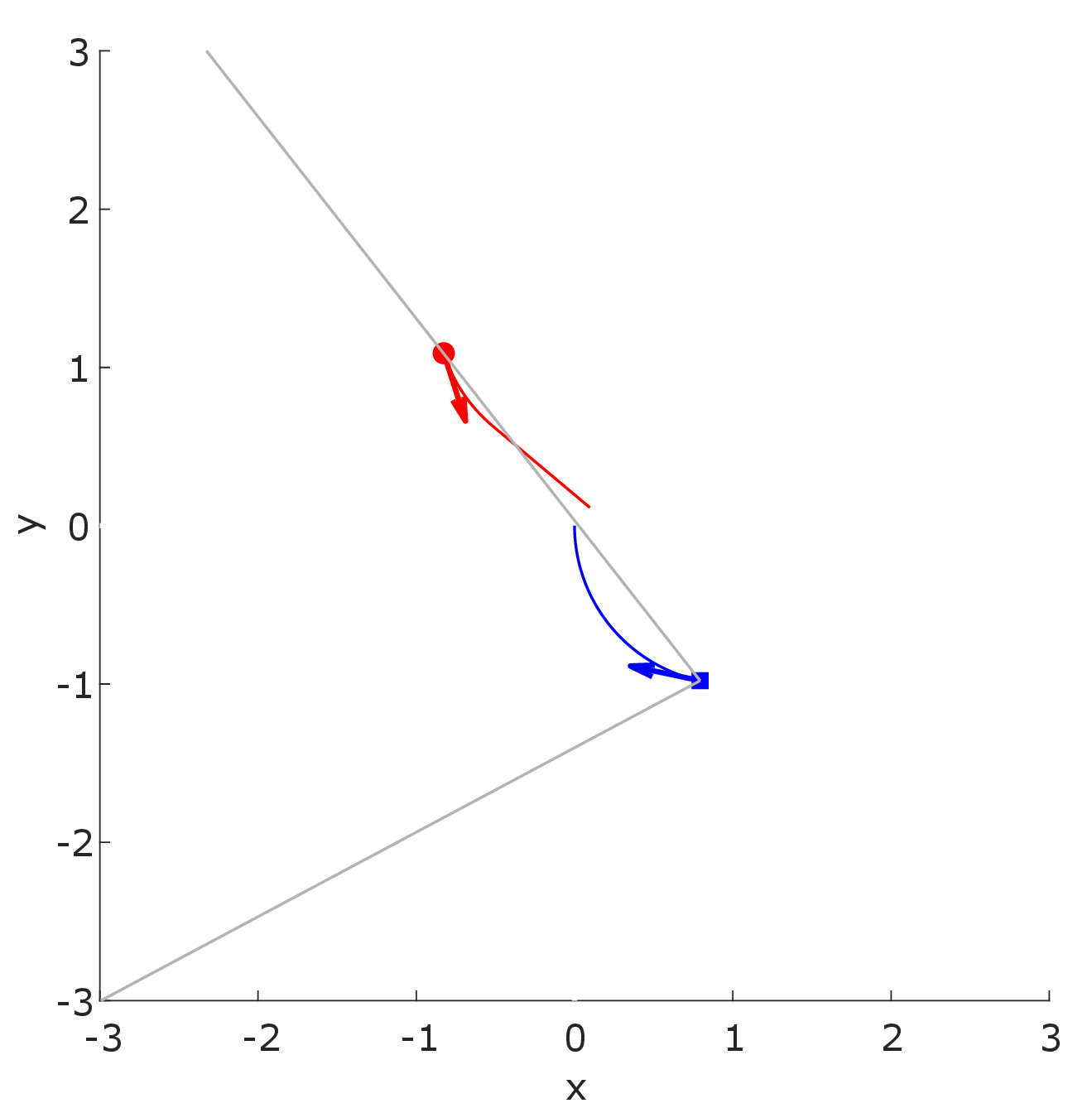}
}
\subfloat[$t=0.5s$ \label{fig:sim3_seq2}]{
\includegraphics[scale=0.28]{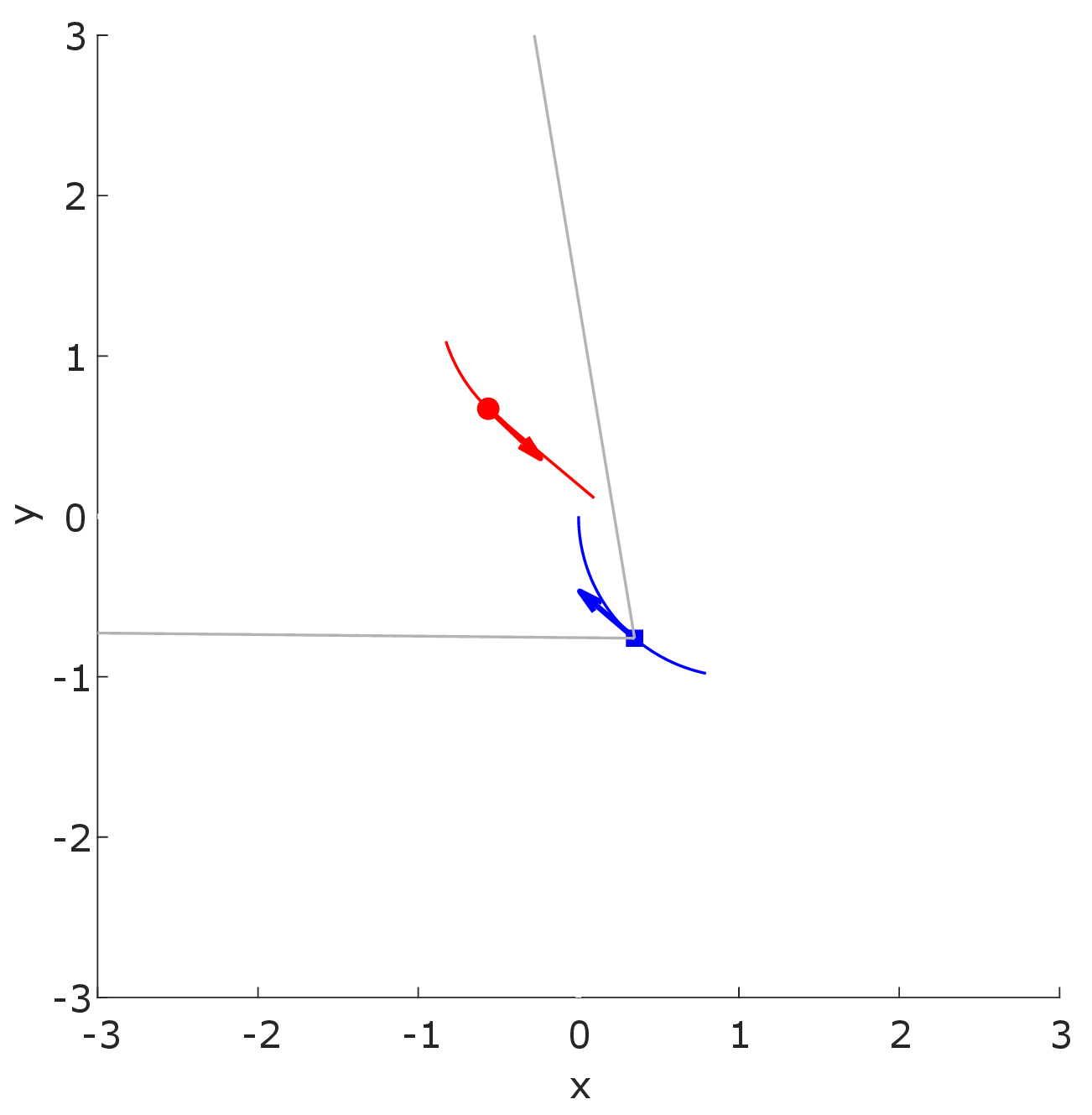}
}\\
\subfloat[$t=1s$ \label{fig:sim3_seq3}]{
\includegraphics[scale=0.28]{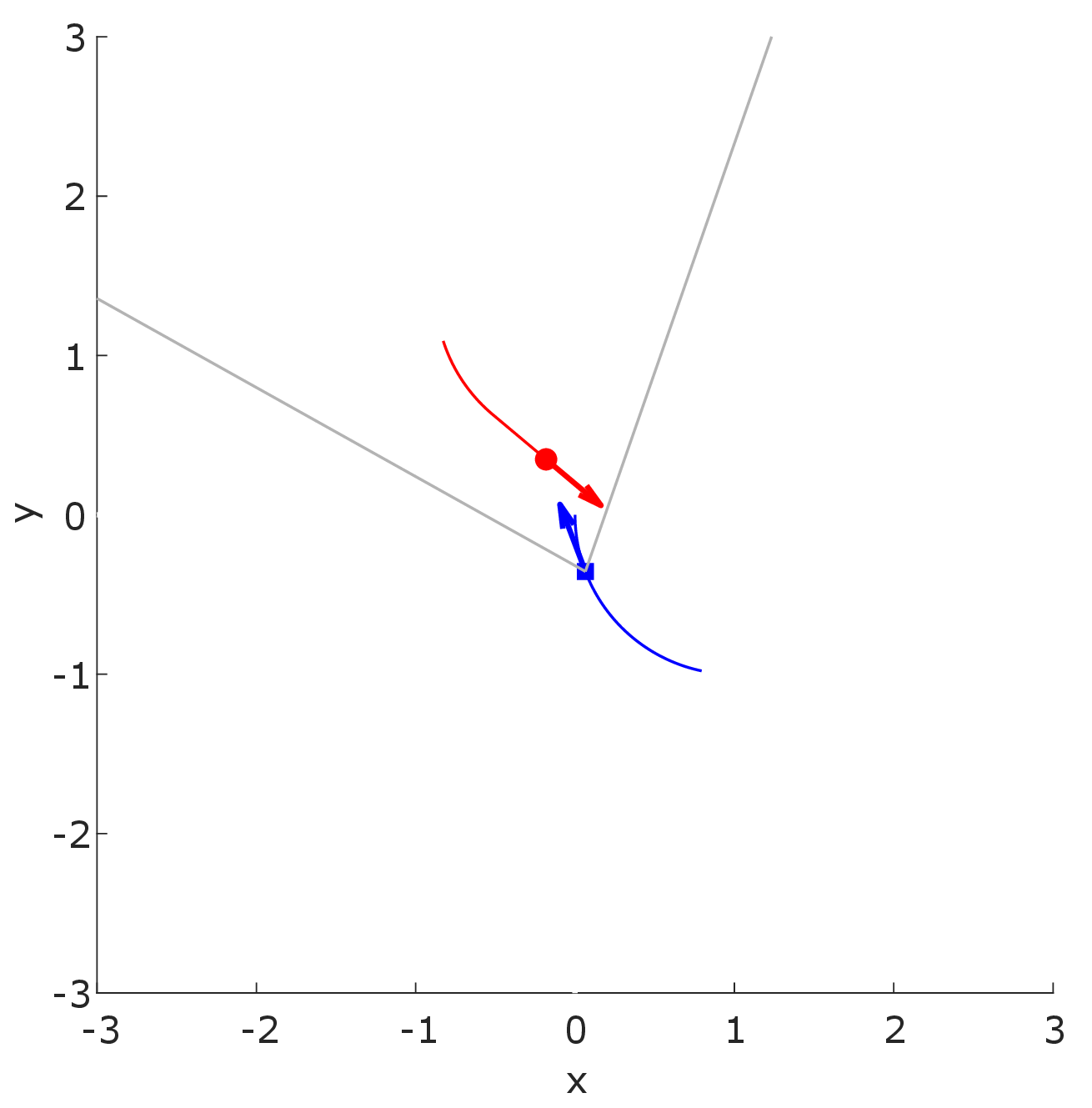}
}
\subfloat[$t=1.36s$ \label{fig:sim3_seq4}]{
\includegraphics[scale=0.28]{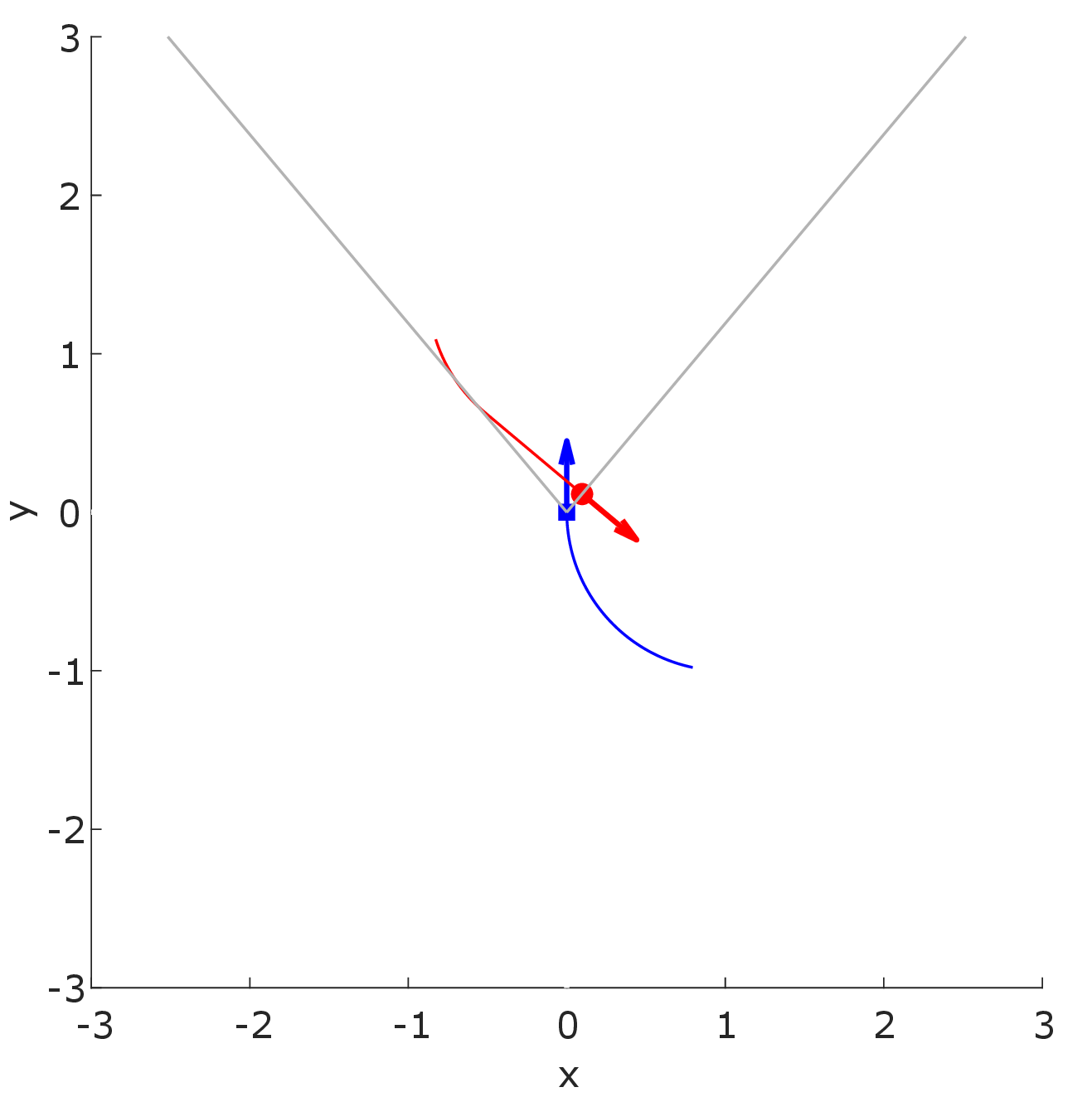}
}\\
\caption{Snapshots of the motion strategies of the players in the realistic space for the fifth simulation. The pursuer is represented by the blue dot, and the evader by the red one. The boundary of the FoV corresponds to the gray lines. The blue curve indicates the entire pursuer's trajectory, and the red one is the full evader's trajectory. The caption of each figure indicates the elapsed time of the simulation when it was taken. \label{fig:sim3snapshots}} 
\end{figure*}

For the fifth simulation, we have the same parameters as the fourth simulation. However, unlike that simulation, the system follows the Universal Surface's alternative tributary trajectory (yellow curve) in the reduced space (see Fig. \ref{fig:sim3_reduced}). For this case, in the realistic space (see Fig. \ref{fig:sim3_realistic}), the evader is again located in the right boundary of the detection region but its heading is pointing towards its interior. The evader aims to escape moving in that direction and rotating counter-clockwise. On the contrary, the pursuer rotates clockwise and pushes away the boundary of the detection region. After some time, similarly to the previous simulation, the evader starts following a straight-line trajectory reaching the right boundary of the detection region, and it escapes. Fig. \ref{fig:sim3snapshots} shows several snapshots of the players' configurations in the realistic space taken during the third simulation.

\section{Conclusions and future work}
\label{sec:conclusions}

In this work, we studied the differential game of keeping surveillance of a Dubins car with an identical Dubins car equipped with a limited field of view sensor, modeled as a semi-infinite cone fixed to its body. The evader wants to escape from the detection region as soon as possible. On the contrary, the pursuer wants to keep surveillance of the evader as much as possible. We found the players' time-optimal motion strategies near the game's end. The analysis of the trajectories reveals the existence of at least two singular surfaces: a Transition Surface and an Evader's Universal Surface. We also found the players' motion strategies and the corresponding trajectories that reach those surfaces. We presented five simulations of the players' motion strategies. 

Additionally, we encountered that the barrier's standard construction produces a surface that partially lies outside the playing space. This suggests that either the evader can escape from all initial positions in the playing space, in which case additional singular surfaces and their corresponding trajectories need to be found to fill the entire playing space or the playing space is bounded by other barrier surfaces that emerge from constructing additional singular surfaces. Unfortunately, in this paper, we cannot formally conclude which of the previous two cases occurs. In particular, the process of discovering additional singular surfaces is challenging since the trajectories found so far are represented by the transcendental equations. An analysis of the behavior of the barrier's trajectories outside the detection region maybe be useful to determine if those trajectories penetrate the detection region further in time and provide insight into missing parts of the barrier's trajectories. We left that study as future work since it may require finding additional singular surfaces. Despite the limitations described above, this paper presents the first study of the proposed pursuit-evasion problem and establishes the foundations for future analysis.

\section*{Declarations}

\begin{itemize}
\item {\bf Funding.} This work was supported by CONAHCYT grant A1-S-21934.
\item {\bf Conflict of interest.} The authors declare that they have no conflict of interest.
\end{itemize}

\bibliography{sn-bibliography}

\end{document}